\documentclass{article}
\usepackage{iclr2027_conference,times}

\usepackage{amsmath,amsfonts,bm}

\def\eqref#1{equation~\ref{#1}}

\def\1{\bm{1}}

\DeclareMathAlphabet{\mathsfit}{\encodingdefault}{\sfdefault}{m}{sl}
\SetMathAlphabet{\mathsfit}{bold}{\encodingdefault}{\sfdefault}{bx}{n}

\usepackage{amsmath,amssymb,booktabs,microtype}
\usepackage{graphicx,float,placeins}
\usepackage{xcolor}
\usepackage{colortbl}
\usepackage{array}
\usepackage{hyperref}
\usepackage{url}
\hypersetup{hidelinks}

\graphicspath{{figures/}}
\definecolor{rankfirstbg}{HTML}{D9EAD3}
\definecolor{ranksecondbg}{HTML}{DCE6F1}
\definecolor{rankthirdbg}{HTML}{FCE5CD}
\definecolor{rankfirstfg}{HTML}{3F7D44}
\definecolor{ranksecondfg}{HTML}{416A8C}
\definecolor{rankthirdfg}{HTML}{A65F24}
\newcommand{\rankfirst}[1]{\cellcolor{rankfirstbg}\textbf{#1}}
\newcommand{\ranksecond}[1]{\cellcolor{ranksecondbg}#1}
\newcommand{\rankthird}[1]{\cellcolor{rankthirdbg}#1}

\title{Beyond Arbitrary Geometry:\\
Topology Generalization in\\
Neural PDE Operators}

\author{Peiyao Chen, Zhouyuan Xu, Ran Ding, Jiansheng Fan, Chen Wang\thanks{Corresponding author: \texttt{chwang@tsinghua.edu.cn}.}\\
Tsinghua University}

\iclrfinalcopy

\begin{document}
\maketitle

\pagestyle{plain}

\begin{abstract}
Neural operators that accept arbitrary meshes are often treated as geometry-general, but unseen domain topology changes both the invariant and decaying subspaces of a PDE operator.
We use Hodge heat flow as a controlled lens on this distinction and introduce TopoBox-3D, where tunnels and cavities vary Betti support while the exact Hodge decomposition separates the harmonic kernel from the positive spectrum.
Across six architectures, models that infer topology implicitly suffer excess matched degradation in 37 of 45 model--task topology-OOD cells, yet cases that change harmonic dimension are not more strongly penalized on average.
The dominant difficulty is instead spectral: the initial Rayleigh quotient is the most stable predictor of error, and spectral broadening adds information for edge and face cochains.
Most strikingly, controlled probes show that explicit incidence and harmonic coordinates do not yield the best kernel-identity accuracy; nevertheless, TNO ranks first in mixed-input nonharmonic accuracy on all six tasks with nontrivial harmonic support.
Together, these results establish topology as a distinct generalization axis beyond arbitrary-geometry compatibility and show that its influence extends across the Hodge spectrum rather than remaining confined to the harmonic kernel.
More broadly, they suggest that global, low-frequency structural priors may help organize predictions in the faster-decaying complementary component, offering a new perspective on how neural operators may generalize across topology as well as geometry.
\end{abstract}

\section{Introduction}

Many-query PDE problems arise in design, uncertainty quantification, inverse problems, and digital twins, where repeatedly invoking a high-fidelity solver is expensive.
Operator learning amortizes this cost by approximating maps between function spaces \citep{chen1995universal,lu2021deeponet,li2021fno,kovachki2023neuraloperator}.
Modern architectures use spectral kernels, multiresolution representations, graphs, neural fields, or attention \citep{cao2021transformer,gupta2021multiwavelet,rahman2023uno,li2023oformer,raonic2024cno,li2024codano}, and increasingly accept point clouds, unstructured meshes, and changing shapes \citep{pfaff2021meshgraphnets,li2023geofno,liu2023nuno,li2023gino,serrano2023coral,hao2023gnot,wu2024transolver,alkin2024upt,mousavi2025rigno,wen2025gaot}.

The phrase \emph{arbitrary geometry}, however, often describes representational compatibility rather than statistical extrapolation.
A model may ingest a non-Cartesian discretization or query arbitrary coordinates without generalizing to a domain type absent from training.
Reference-domain methods make this distinction explicit: Geo-FNO learns a deformation to a regular latent grid \citep{li2023geofno}, while DIMON and Diffeomorphism Neural Operator transport problems to shared templates \citep{yin2024dimon,zhao2025dno}.
These constructions are effective within diffeomorphic families, but a diffeomorphism cannot create a tunnel or enclosed cavity.
Masks, direct Fourier evaluation, graphs, and neural fields permit topology to vary \citep{liu2023dafno,lingsch2023ndft,li2023gino,serrano2023coral}; their standard evaluations nevertheless rarely isolate whether test topology lies outside the training support.

We stratify a domain family by its Betti vector $\boldsymbol\beta(\Omega)$.
A test case is \emph{fixed-topology geometry-OOD} when shape or pose is extrapolative but Betti support matches training, and \emph{topology-OOD} when $\boldsymbol\beta(\Omega_{\rm test})$ lies outside that support.
This is deliberately a homological definition: Betti numbers do not classify domains completely.
Prior work has studied evolving cracks \citep{liu2023dafno}, local-to-global transfer on new domains \citep{huang2026domain}, and surface PDEs across topology-changing shapes \citep{welschinger2026shapepde}; our aim is not to claim that topology-changing PDE learning is unprecedented, but to isolate topology support and connect it to an operator-level mechanism.

For the $k$-form Hodge Laplacian under homogeneous absolute boundary conditions,
\begin{equation}
\ker\Delta_k\cong H^k(\Omega),\qquad
S_T=e^{-\kappa T\Delta_k}=P_H+e^{-\kappa T\Delta_{k,\perp}}P_\perp,
\label{eq:hodge-split}
\end{equation}
where $P_H$ is the mass-orthogonal harmonic projector,
$P_\perp=I-P_H$, $\Delta_{k,\perp}$ is the restriction of $\Delta_k$ to the positive spectral subspace, and $\dim\ker\Delta_k=\beta_k(\Omega)$.
Topology can therefore alter the dimension and orientation of the invariant harmonic space, whereas geometry and discretization reshape the positive spectrum.
Compatible FEEC discretizations preserve this relation \citep{arnold2010feec,arnold2017parabolic}, turning topology generalization into a test of kernel recognition, subspace invariance, and positive-spectrum filtering.

We use \textbf{TopoBox-3D} to extend the study of cross-domain neural operators from arbitrary geometry to unseen topology.
Independent tunnel and cavity control, $k=0,1,2$ Hodge heat, and matched geometry/topology protocols let us ask whether a model that accepts changing meshes can extrapolate beyond observed Betti support, and where any failure appears in the Hodge decomposition across six architecture families.
Unseen Betti support usually adds degradation when topology must be inferred implicitly, but harmonic-dimension changes do not amplify that penalty, and the most stable error driver is the positive-spectrum frequency excited by the input.
Moreover, controlled seed-0 probes show that the model with explicit incidence and harmonic access is not the best at reproducing harmonic identity, yet it has the lowest mixed-input nonharmonic error on every task with nontrivial harmonic support.
The contribution is therefore a new generalization perspective and the operator-level evidence it enables: topology is distinct from geometry, and its effect spans kernel recognition, cross-subspace leakage, and complementary-spectrum prediction, while explicit kernel information may still organize how the nonharmonic component is learned.

\section{Related Work and Background}

\paragraph{Neural operators and irregular-geometry representations.}
DeepONet and FNO established complementary branch--trunk and spectral constructions \citep{lu2021deeponet,li2021fno}; subsequent work introduced multiwavelet, U-shaped, convolutional, and attention operators \citep{gupta2021multiwavelet,rahman2023uno,raonic2024cno,cao2021transformer,li2023oformer}.
To operate beyond regular grids, neural operators have adopted learned coordinate maps \citep{li2023geofno}, domain indicators \citep{liu2023dafno}, nonuniform Fourier transforms \citep{lingsch2023ndft}, spatial partitions \citep{liu2023nuno}, graph--spectral encoders \citep{li2023gino}, and implicit neural fields \citep{serrano2023coral}.
MeshGraphNets and related PDE solvers propagate directly on simulation meshes \citep{pfaff2021meshgraphnets,brandstetter2022messagepassing}; GNOT and Transolver use geometric gating or physics tokens \citep{hao2023gnot,wu2024transolver}; UPT compresses variable samples to a fixed latent set \citep{alkin2024upt}; and RIGNO, GAOT, and Transolver++ improve accuracy and scalability on irregular domains \citep{mousavi2025rigno,wen2025gaot,luo2025transolverpp}.
These methods substantially broaden the geometries and discretizations that a model can ingest, but input compatibility alone does not specify how it extrapolates across geometry families absent from training.

\paragraph{Operators for cross-geometry generalization.}
A complementary line of work explicitly targets transfer between domains.
Reference-domain approaches align geometries through learned or prescribed coordinate transformations, including Geo-FNO, DIMON, and Diffeomorphism Neural Operator \citep{li2023geofno,yin2024dimon,zhao2025dno}; domain-decomposition methods instead assemble learned local solution operators on new domains \citep{huang2026domain}.
Other studies address changing cracks \citep{liu2023dafno} or surface PDEs across shape and topology variation \citep{welschinger2026shapepde}.
Within this broader cross-geometry direction, geometric deep learning provides relational and symmetry-aware representations \citep{bronstein2021geometric}, while higher-order networks encode oriented simplices, incidence relations, equivariance, and topological summaries \citep{battiston2020higherorder,roddenberry2021scone,bodnar2021mpsn,eijkelboom2023empsn,verma2024topnets}.
Discrete Hodge operators expose diffusion and harmonic structure on such complexes \citep{schaub2020randomwalks}, and recent Hodge Spectral Duality and Topological Neural Operators incorporate related structure into learned PDE operators \citep{zheng2026hsd,bastian2026tno}.
We treat these as structured approaches to cross-geometry generalization rather than as a separate problem class: our contribution is not a new Hodge architecture, but a controlled comparison of extrapolation beyond observed Betti support and an audit of whether learned maps preserve the kernel identities in Equation~\ref{eq:hodge-split}.

\section{TopoBox-3D Benchmark}
\label{sec:benchmark}

\begin{figure}[H]
    \centering
    \includegraphics[width=\linewidth]{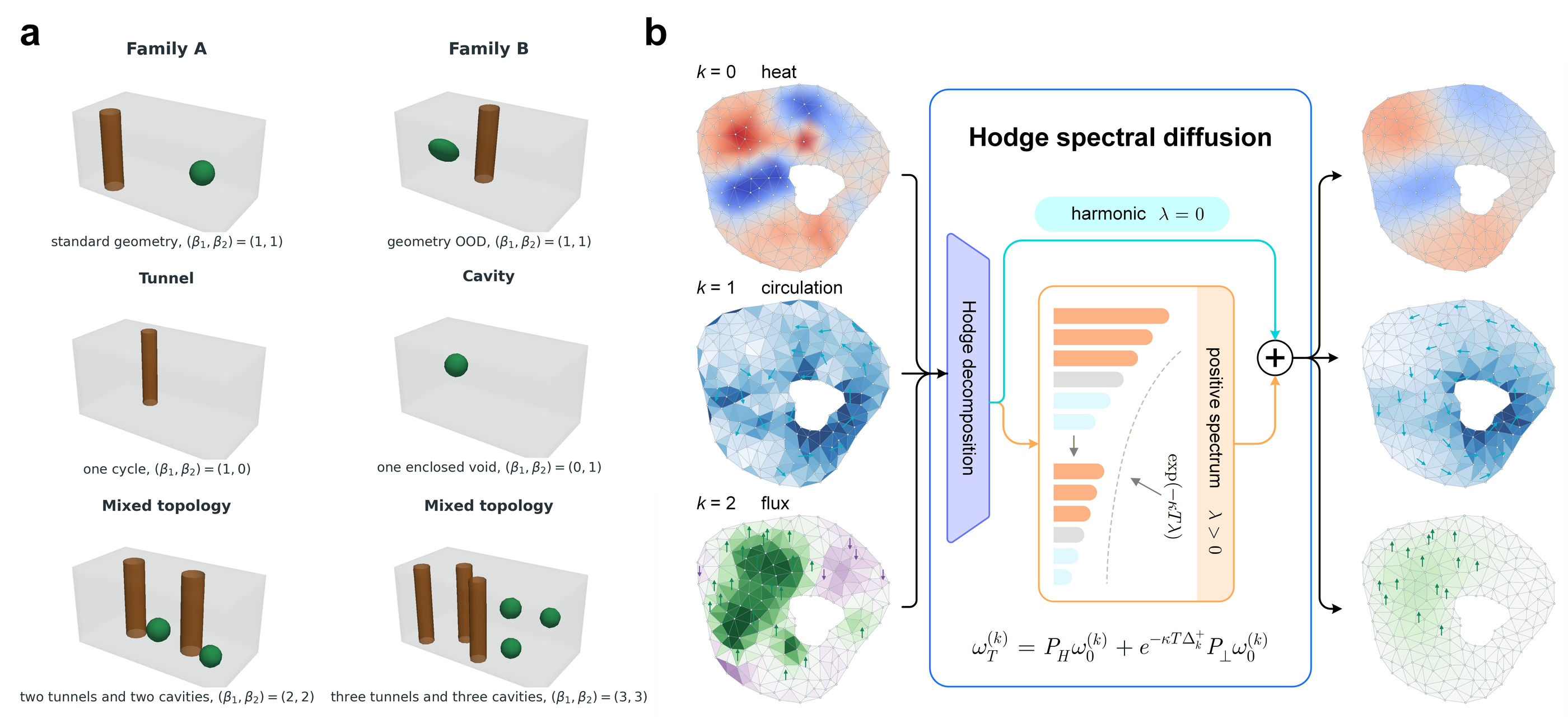}
    \caption{\textbf{TopoBox-3D geometry and Hodge diffusion.}
    (a) Representative domain families. Family A and Family B share $(\beta_1,\beta_2)=(1,1)$ but differ geometrically, while tunnel-only, cavity-only, and mixed cases independently vary cycles and enclosed voids.
    (b) Schematic physical interpretations of Hodge diffusion for $k=0,1,2$, corresponding to scalar, circulation, and flux fields.}
    \label{fig:topobox-overview}
\end{figure}

TopoBox-3D separates ordinary geometric extrapolation from extrapolation across homological support (Figure~\ref{fig:topobox-overview}).
Let $D=[0,2]\times[0,1]^2$ and
\begin{equation}
\Omega=D\setminus\left(\bigcup_{i=1}^{g}\mathcal T_i\cup\bigcup_{j=1}^{c}\mathcal C_j\right),
\label{eq:topobox-domain}
\end{equation}
where $\mathcal T_i$ is a through-tunnel and $\mathcal C_j$ a closed cavity.
Voids are disjoint, cavities remain separated from the exterior, and the material domain remains connected.
The standard family uses $z$-aligned circular tunnels and spherical cavities; the geometry-OOD family uses tilted tunnels and rotated ellipsoidal cavities.
Under the enforced clearances, $\boldsymbol\beta(\Omega)=(1,g,c,0)$.
After tetrahedralization we reconstruct the oriented chain complex, recompute ranks and Euler characteristic, and verify $B_1B_2=B_2B_3=0$, so topology is measured on the numerical domain rather than accepted from generator labels.

Each mesh defines $C^0(K)\xrightarrow{B_1^\top}C^1(K)\xrightarrow{B_2^\top}C^2(K)\xrightarrow{B_3^\top}C^3(K)$.
For $k\in\{0,1,2\}$ we solve
\begin{equation}
\partial_t\omega^{(k)}+\kappa\Delta_k\omega^{(k)}=0,
\qquad \omega_T^{(k)}=e^{-\kappa T\Delta_k}\omega_0^{(k)},
\label{eq:benchmark-heat}
\end{equation}
with diagonal mass matrices, homogeneous absolute boundary conditions, $\kappa=1$, $T=0.1$, and 100 Crank--Nicolson steps.
The three degrees provide complementary physical analogues: $k=0$ gives scalar heat-like conduction, $k=1$ describes diffusive transport of velocity or circulation on oriented edges, and $k=2$ describes diffusion of face-based fluxes such as magnetic flux.
Together they probe scalar, circulation, and flux dynamics within one operator family rather than a single physical field type.
Figure~\ref{fig:topobox-overview}(b) schematically illustrates the scalar, circulation, and flux diffusion associated with $k=0,1,2$.
The $k=2$ operator retains its required $B_3$ contribution.
Each geometry and degree receives four exact/coexact/harmonic energy mixtures: $(1/2,1/2,0)$, $(4/9,4/9,1/9)$, $(1/3,1/3,1/3)$, and $(1/4,1/4,1/2)$.
Unavailable harmonic energy is redistributed over realizable components; $k=0$ uses smooth mean-zero fields as a topology-insensitive control.

Table~\ref{tab:protocols} summarizes the four generalization protocols.
Protocol A holds Betti support fixed while changing the geometry family; Protocols B--D extrapolate beyond the observed support of tunnels, cavities, or both.
Each protocol contains 800 training, 120 validation, 200 Test-IID, and 200 Test-OOD geometries, giving 5,280 domains and 63,360 PDE instances.
Geometry ID is the atomic split unit, and Test-OOD is never used for model selection.

\begin{table}[H]
\caption{TopoBox-3D protocols. In-support topologies apply to training, validation, and Test-IID.}
\label{tab:protocols}
\centering
\scriptsize
\begin{tabular}{@{}c p{0.29\linewidth} p{0.17\linewidth} p{0.34\linewidth}@{}}
\toprule
Protocol & In-support topologies & Test-OOD topology & Generalization shift \\
\midrule
A & $(\beta_1,\beta_2)=(1,1)$, family A
  & $(1,1)$, family B
  & fixed-topology geometry shift \\
B & $\beta_1\in\{0,1,2\},\ \beta_2=0$
  & $(3,0)$
  & unseen tunnel support ($\beta_1$-OOD) \\
C & $\beta_1=0,\ \beta_2\in\{0,1,2\}$
  & $(0,3)$
  & unseen cavity support ($\beta_2$-OOD) \\
D & $(\beta_1,\beta_2)\in\{0,1,2\}^2$
  & $(3,3)$
  & mixed topology-OOD \\
\bottomrule
\end{tabular}
\end{table}

\section{Experimental Setup}
\label{sec:setup}

We compare six approximately capacity-matched models spanning four levels of structural access (Table~\ref{tab:models}).
This is not a single-factor architecture ablation: processors, tokenization, and native inputs also differ.
We therefore emphasize within-model IID-to-OOD degradation and structure-access strata rather than treating ranks as causal effects of one feature.

The access strata describe information available to the processor, not merely the file format of its input.
TNO receives the oriented incidence maps across degrees and an $M_k$-orthonormal basis for the active harmonic subspace \citep{bastian2026tno}.
MGN operates on the original same-degree simplex adjacency, preserving mesh connectivity but not cross-degree chain structure \citep{pfaff2021meshgraphnets}.
RIGNO and GAOT construct regional or latent graphs from coordinates \citep{mousavi2025rigno,wen2025gaot}, while Transolver and GNOT interact through coordinate-indexed attention tokens \citep{wu2024transolver,hao2023gnot}.
All adapters predict one scalar canonical cochain value per active vertex, edge, or face; no method is evaluated through a lossy cochain-to-vertex conversion.
Coordinates, local measures, boundary indicators, and signed-distance features are shared wherever they are native to the corresponding adapter, while Betti numbers and harmonic bases are withheld from the five implicit-topology models.

\begin{table}[H]
\caption{Models and effective access to TopoBox structure. Parameter ranges cover $k=0$ versus $k=1,2$.}
\label{tab:models}
\centering
\scriptsize
\resizebox{\linewidth}{!}{%
\begin{tabular}{@{}lllr@{}}
\toprule
Model & Processor & Structural access & Parameters \\
\midrule
MGN\textsuperscript{1} & 6-layer MPNN & same-degree simplex adjacency & 1.162M \\
RIGNO & regional graph, 7 steps & coordinate-reconstructed graph & 1.197--1.198M \\
Transolver & 6-layer physics attention & coordinate tokens, local geometry & 1.210--1.212M \\
GNOT & 4-layer normalized attention & coordinate tokens, local geometry & 1.135--1.136M \\
GAOT & MAGNO + transformer & coordinate latent graph & 1.186M \\
TNO\textsuperscript{2} & 4-layer cross-degree operator & $B_1,B_2,B_3$ + harmonic basis & 1.168M \\
\bottomrule
\end{tabular}}
\end{table}
\footnotetext[1]{MGN is our degree-specific MeshGraphNets-style adaptation, operating on same-degree simplices rather than the original rollout simulator.}
\footnotetext[2]{TNO is our implementation of the published cross-degree incidence and harmonic mechanism because no official code was available.}
\setcounter{footnote}{2}

All models train for 300 epochs with AdamW, initial learning rate $2\times10^{-4}$, weight decay $10^{-4}$, cosine decay to $10^{-5}$, gradient clipping at 1, batch size 1, and seeds $\{0,1,2\}$, totaling $6\times4\times3\times3=216$ runs.
Validation is restricted to training topology support; the best validation checkpoint is evaluated once on Test-IID and Test-OOD.
Training minimizes mass-weighted relative MSE; reporting uses the corresponding mass-weighted relative $L^2$ error $E$.
For a prediction $\widehat w$ and target $w$, these are
\begin{equation}
\mathcal L_{\rm train}=\frac{\|\widehat w-w\|_{M_k}^{2}}{\|w\|_{M_k}^{2}+\epsilon},
\qquad
E=\frac{\|\widehat w-w\|_{M_k}}{\|w\|_{M_k}+\epsilon},
\label{eq:losses}
\end{equation}
with the same diagonal mass matrix used by the discrete PDE and $\epsilon=10^{-8}$, matching the released evaluation code.
For model $m$, protocol $p$, and degree $k$, define
\begin{equation}
R_{m,p,k}=E^{\rm OOD}_{m,p,k}/E^{\rm IID}_{m,p,k},\qquad
Q_{m,p,k}=R_{m,p,k}/R_{m,A,k}.
\label{eq:rq}
\end{equation}
Ratios are formed within seed before taking the three-seed median.
Because four initial conditions share each geometry, uncertainty is obtained by independently resampling IID and OOD geometry IDs.
Validation is evaluated every five epochs, early stopping is disabled, and every run reaches epoch 300; the best in-support validation checkpoint is selected before either test split is inspected.
Ranks are computed separately within each of the 12 protocol--degree tasks and then averaged, so a low-error scalar task cannot dominate a higher-order task solely through its numerical scale.
Geometry-cluster bootstrap confidence intervals use 1,000 independent resamples of the 200 IID and 200 OOD geometry IDs and retain all four initial conditions whenever a geometry is selected.

\section{Results}
\label{sec:results}

\subsection{Predictive performance}

Direct access to the original discrete structure is associated with stronger absolute accuracy, although not monotonically and not as a single-factor causal effect.
Across 12 tasks, TNO has mean IID/OOD ranks 1.58/1.67 and ranks in the top three everywhere; MGN wins 5/12 IID and 5/12 OOD tasks and is first for every $k=0$ protocol and seed.
Together they account for 11/12 task wins in both IID and OOD evaluation.
The pattern is degree dependent: raw vertex adjacency closely matches scalar diffusion, whereas TNO more often leads for edge and face cochains.
Coordinate-attention and latent-graph models show task specialization rather than a common ordering.
Low $R$ alone is not sufficient evidence of robustness: GAOT obtains favorable ratios in several cells from a high IID error floor, while RIGNO provides a more competitive accuracy--stability tradeoff.

Tables~\ref{tab:iid-loss} and~\ref{tab:ood-loss} report the complete Test-IID and Test-OOD errors underlying these ranks.
Column labels concatenate protocol and cochain degree: for example, A0 denotes Protocol A at $k=0$, and the same convention extends through D2.

\begin{table}[H]
\centering\scriptsize
\caption{Median Test-IID mass-weighted relative $L^2$ error over three seeds. \textcolor{rankfirstfg}{\textbf{Rank 1}}, \textcolor{ranksecondfg}{\textbf{Rank 2}}, and \textcolor{rankthirdfg}{\textbf{Rank 3}} indicate the three lowest errors in each column.}
\label{tab:iid-loss}
\resizebox{\linewidth}{!}{\begin{tabular}{@{}lrrrrrrrrrrrr@{}}
\toprule
& A0&A1&A2&B0&B1&B2&C0&C1&C2&D0&D1&D2\\\midrule
MGN&\rankfirst{.0101}&\ranksecond{.2191}&\ranksecond{.3487}&\rankfirst{.0091}&\rankthird{.2254}&\rankfirst{.2756}&\rankfirst{.0086}&.1725&\ranksecond{.3431}&\rankfirst{.0103}&\rankthird{.2480}&\ranksecond{.3603}\\
RIGNO&.0258&\rankthird{.2646}&\rankthird{.3491}&.0232&.2333&\ranksecond{.3278}&.0228&\rankthird{.1651}&\rankthird{.3725}&.0262&.2524&\rankthird{.3994}\\
Transolver&.0354&.4188&.5034&.0347&.3837&.4399&.0317&.2755&.4917&.0351&.3655&.5730\\
GNOT&\rankthird{.0202}&.3257&.4171&\rankthird{.0183}&\ranksecond{.2161}&.4137&\rankthird{.0161}&\rankfirst{.0879}&.4211&\rankthird{.0221}&\ranksecond{.2296}&.4942\\
GAOT&.3190&.5408&.8479&.3199&.5413&.5166&.3253&.5174&.6656&.3178&.5476&.7098\\
TNO&\ranksecond{.0155}&\rankfirst{.1029}&\rankfirst{.1760}&\ranksecond{.0144}&\rankfirst{.1216}&\rankthird{.3848}&\ranksecond{.0131}&\ranksecond{.1622}&\rankfirst{.1840}&\ranksecond{.0161}&\rankfirst{.1362}&\rankfirst{.2478}\\
\bottomrule\end{tabular}}
\end{table}

\begin{table}[H]
\centering\scriptsize
\caption{Median Test-OOD mass-weighted relative $L^2$ error over three seeds. \textcolor{rankfirstfg}{\textbf{Rank 1}}, \textcolor{ranksecondfg}{\textbf{Rank 2}}, and \textcolor{rankthirdfg}{\textbf{Rank 3}} indicate the three lowest errors in each column.}
\label{tab:ood-loss}
\resizebox{\linewidth}{!}{\begin{tabular}{@{}lrrrrrrrrrrrr@{}}
\toprule
& A0&A1&A2&B0&B1&B2&C0&C1&C2&D0&D1&D2\\\midrule
MGN&\rankfirst{.0105}&\ranksecond{.2288}&\rankthird{.4760}&\rankfirst{.0138}&\rankthird{.2699}&\rankfirst{.3486}&\rankfirst{.0100}&.1986&\rankthird{.4633}&\rankfirst{.0134}&.2864&\ranksecond{.4088}\\
RIGNO&.0265&\rankthird{.2651}&\ranksecond{.4071}&\rankthird{.0292}&\ranksecond{.2428}&\ranksecond{.3971}&.0234&\ranksecond{.1752}&\ranksecond{.4568}&.0291&\ranksecond{.2696}&\rankthird{.4681}\\
Transolver&.0379&.4160&.5933&.0493&.4230&.5789&.0336&.2823&.6522&.0406&.3863&.6555\\
GNOT&\rankthird{.0221}&.3294&.4864&.0350&.3079&.5599&\rankthird{.0196}&\rankfirst{.1056}&.5769&\rankthird{.0289}&\rankthird{.2864}&.6239\\
GAOT&.3205&.5297&.8581&.3155&.5359&.6856&.3236&.5216&.7928&.3140&.5413&.8253\\
TNO&\ranksecond{.0161}&\rankfirst{.1086}&\rankfirst{.1770}&\ranksecond{.0216}&\rankfirst{.1058}&\rankthird{.5210}&\ranksecond{.0160}&\rankthird{.1881}&\rankfirst{.1428}&\ranksecond{.0196}&\rankfirst{.1169}&\rankfirst{.2218}\\
\bottomrule\end{tabular}}
\end{table}

Together, the two tables expose three patterns that are obscured by a single aggregate score.
First, the $k=0$ problems form a distinct regime: MGN has the lowest error in every protocol, consistent with direct vertex adjacency being well matched to scalar diffusion.
Second, TNO is strongest on most $k=1,2$ tasks, including the difficult mixed-topology D-$k=2$ case, but is not uniformly dominant; its B-$k=2$ OOD error is 0.5210 versus 0.3486 for MGN.
Third, the error floor differs substantially across processors.
For example, GAOT's ratios can remain near one while both IID and OOD errors are large, whereas GNOT is unusually strong on C-$k=1$ but degrades sharply under topology shifts that affect the active harmonic degree.
These observations establish the absolute-error context needed to interpret the degradation ratios $R$ and matched penalties $Q$ analyzed next.

\subsection{Geometry versus topology generalization}
\label{sec:geometry-topology}

\begin{figure}[H]
\centering
\includegraphics[width=\linewidth]{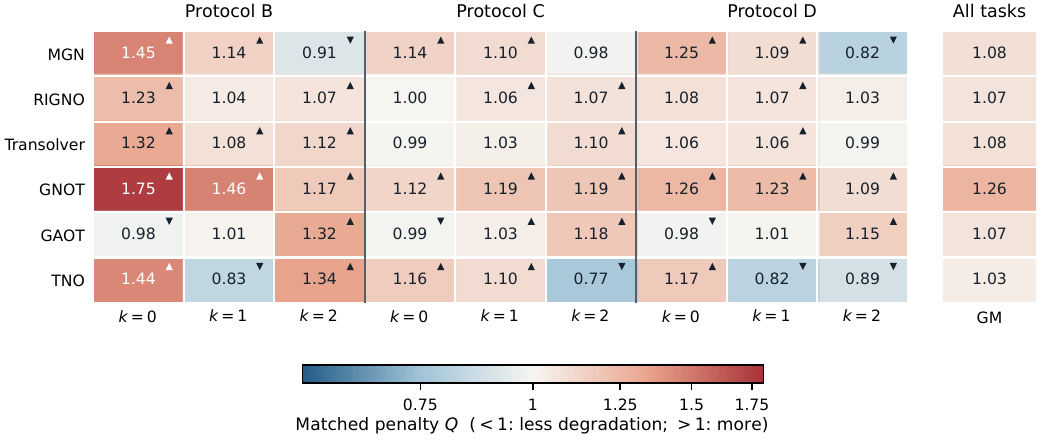}
\caption{\textbf{Matched topology penalty across all topology-OOD tasks.}
Each cell reports $Q=R_p/R_A$ for protocols B--D and degrees $k=0,1,2$; blue ($Q<1$) indicates less degradation than the fixed-topology Protocol A baseline and red ($Q>1$) indicates more.
Up/down triangles mark 95\% geometry-cluster intervals entirely above/below one; the right column gives the geometric mean across all nine tasks.}
\label{fig:performance-topology}
\end{figure}

Figure~\ref{fig:performance-topology} reports the matched penalty across all B/C/D tasks.
Across the 45 model--degree comparisons for the five models without the full chain complex and harmonic basis, 37 (82.2\%) have $Q>1$, with geometric mean $Q=1.110$ (Table~\ref{tab:aggregate-topology-penalty}).
Topology-OOD degradation is therefore widespread across protocols and degrees rather than confined to the tasks whose harmonic dimension changes.

The same pattern holds, but is not amplified, in the 20 implicit-model cells whose harmonic dimension changes: 17/20 have $Q>1$, with geometric mean $Q=1.083$, and 13 geometry-cluster bootstrap intervals lie entirely above one.
Their mean penalty is slightly smaller than the all-degree mean, indicating that topology-OOD error is not localized to the harmonic kernel.
All four TNO comparisons instead have $Q<1$ (geometric mean 0.827).
Because harmonic-dimension changes do not amplify $Q$ in the broader model set, this reversal suggests that the benefit of TNO's harmonic prior is not confined to fitting the kernel subspace, but extends to generalization on the complementary spectrum.

The matched ratio is a stress-test contrast, not a causal estimate of a Betti-number effect.
Its fixed-topology baseline reduces sensitivity to scale differences across $k$ and model-specific IID calibration.
Unseen Betti support changes the kernel dimension, but protocols B--D also change boundary count, mesh size, and positive spectrum.
Because these factors co-vary, we next ask which geometry, topology, and spectral descriptors consistently predict relative error across models and tasks.

\subsection{Error drivers and spectral broadening}

The matched penalties in Section~\ref{sec:geometry-topology} show where generalization deteriorates, but not which sample properties make prediction difficult.
Because protocols B--D jointly change topology, discretization, and spectrum, we correlate the reported sample-level relative $L^2$ error with seven geometry, topology, and spectral descriptors after taking the median over seeds.
The initial Rayleigh quotient $\mathcal R(w_0)=w_0^\top K_kw_0/(w_0^\top M_kw_0)$ is the most stable predictor: its median Spearman correlation across 72 model--task cells is 0.784, and it gives the largest single-variable $R^2$ in 43/72 cells.
This dominance identifies the mean positive-spectrum frequency excited by the input as the main cross-task difficulty axis: larger $\mathcal R(w_0)$ places more energy in rapidly decaying modes and leaves a weaker final-time signal to reconstruct.
Consistently, target retention $q_T=\|w_T\|_M/\|w_0\|_M$ is strongly negatively associated with error, but it uses the target norm and also enters the relative-error denominator, so it is a predictive summary rather than an independent cause.

Harmonic fraction $\rho_H$ is protective for higher-order cochains, consistent with the exact preservation of zero-eigenvalue energy.
Among the positively associated structural descriptors, token count $\log n_k$ measures discretization burden, while $\beta_k$ and $\beta_1+\beta_2$ summarize degree-specific and total topological complexity; all three are weak globally but become substantially more informative at $k=2$.
The smallest positive eigenvalue $\log\lambda_1^+$ is the weakest individual predictor overall, yet it too strengthens at $k=2$.
This degree dependence helps interpret Figure~\ref{fig:performance-topology}: topology-OOD penalties need not be localized to changes in kernel dimension because the accompanying discretization and positive-spectrum shifts can also change difficulty.
The descriptors nevertheless co-vary, so we fit each separately with geometry-clustered uncertainty and interpret cross-model/task stability as association rather than mechanism.
Under this conservative comparison, $\mathcal R(w_0)$ remains dominant and, unlike $q_T$, is available from the input alone.

The Rayleigh quotient summarizes spectral location but not spectral width: two inputs can have the same mean eigenvalue while distributing energy very differently across modes.
For normalized spectral weights $p_i$ over the full nonnegative spectrum, including harmonic zero modes, the exact heat semigroup satisfies
\begin{equation}
J_T=\log q_T+\kappa T\mathcal R(w_0)\geq0,
\label{eq:jensen}
\end{equation}
which is zero when the energy is supported on a single eigenvalue and whose leading small-time term is proportional to the spectral variance.
Thus $J_T$ measures broadening beyond the Rayleigh mean.

\begin{figure}[H]
\centering
\includegraphics[width=\linewidth]{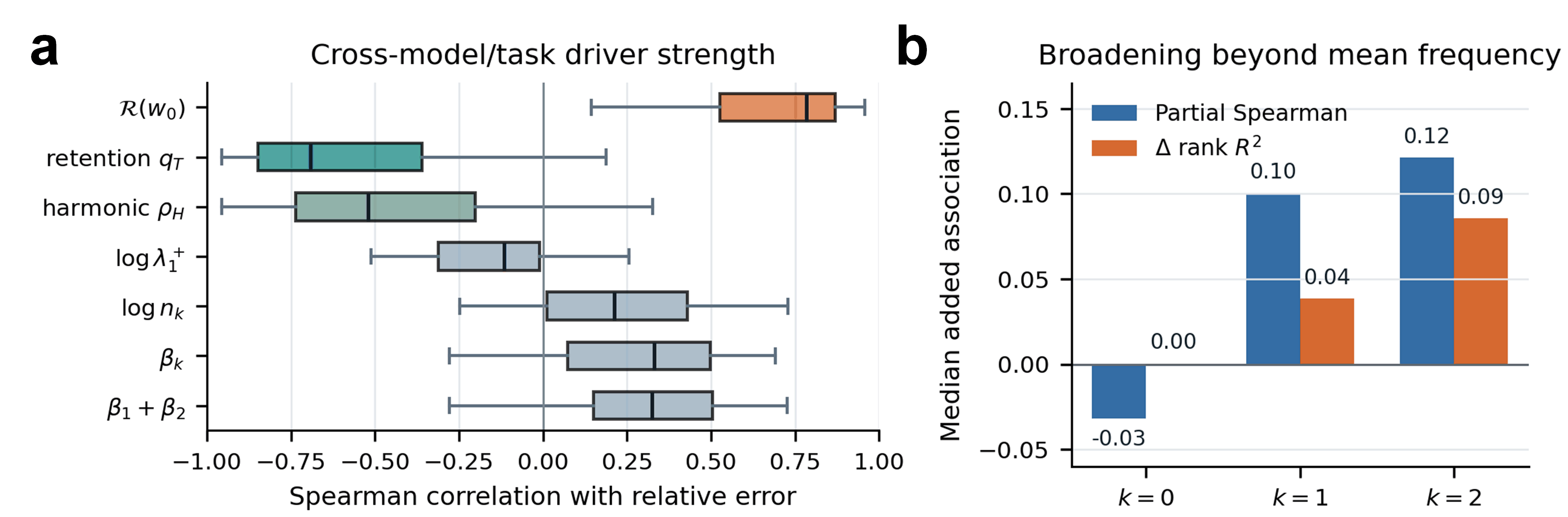}
\caption{\textbf{Drivers of relative error.} (a) Distributions of univariate Spearman correlations over model--task cells. (b) Information added by spectral broadening $J_T$ after controlling for the Rayleigh mean: partial Spearman correlation and the increment in rank-regression $R^2$, summarized across model--task cells at each degree.}
\label{fig:spectral-drivers}
\end{figure}

Figure~\ref{fig:spectral-drivers}(b) asks whether this width information still predicts error after controlling for $\mathcal R(w_0)$.
Although the marginal correlation of $J_T$ is 0.503, its overall median partial correlation is only 0.009 and its median incremental rank-$R^2$ is 0.010, so most of the apparent association is already explained by mean frequency.
The degree-stratified result is more informative: the partial correlations are $-0.032$, 0.100, and 0.121, while the incremental $R^2$ values are 0.0003, 0.038, and 0.086 for $k=0,1,2$.
Spectral broadening therefore adds essentially no information for scalar diffusion but becomes a modest, increasingly useful correction for edge and face cochains; mean spectral decay remains the primary difficulty axis.

\subsection{Topology priors extend beyond the kernel}
\label{sec:topology-audit}

Equation~\ref{eq:hodge-split} motivates three controlled probe axes:
$I_H=\|\widehat S_TP_H-P_H\|$, $L_{H\to\perp}=\|P_\perp\widehat S_TP_H\|$, and $L_{\perp\to H}=\|P_H\widehat S_TP_\perp\|$.
For conventional mixed inputs, we separately report the complementary-component relative error $E_\perp^{\mathrm{mix}}=\|P_\perp(\widehat w_T-w_T)\|_M/\|P_\perp w_T\|_M$ (mass-normalized forms and sampling scopes are in Appendix~\ref{app:math}).
Thus $I_H$ measures empirical identity accuracy within the kernel, the leakage terms measure cross-subspace coupling, and $E_\perp^{\mathrm{mix}}$ measures nonharmonic output accuracy under mixed inputs; it is not a pure-nonharmonic probe error.

\begin{figure}[H]
\centering
\includegraphics[width=\linewidth]{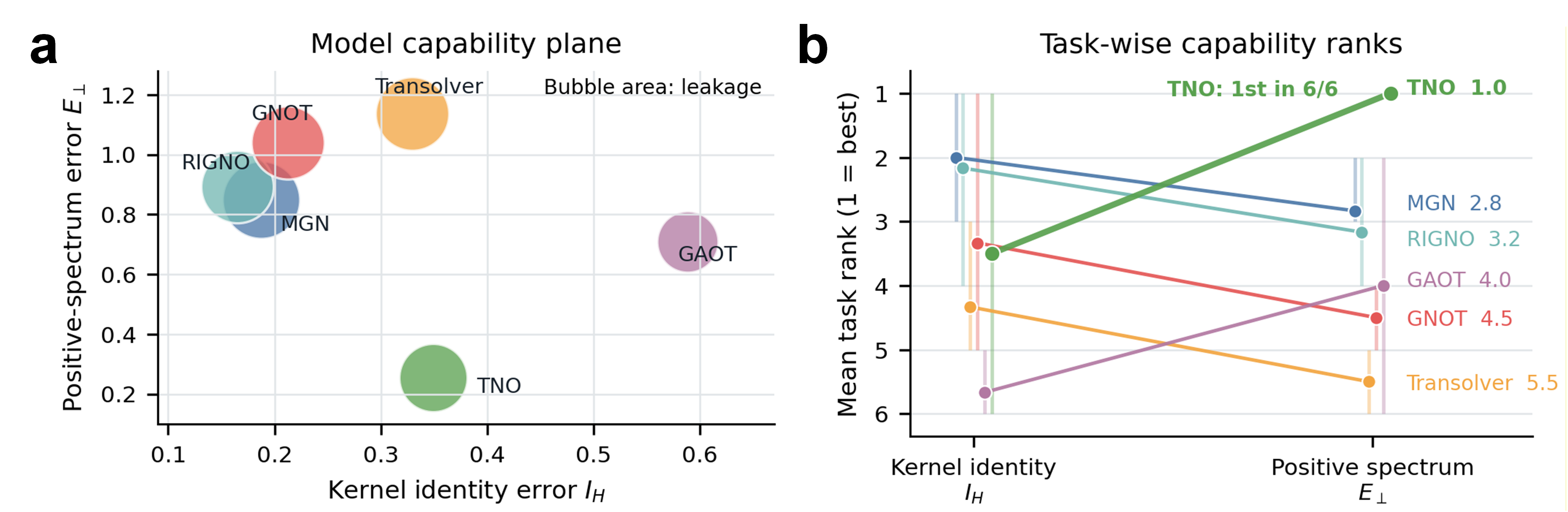}
\caption{\textbf{Architecture-dependent capability asymmetry.} (a) Controlled seed-0 kernel-identity error versus mixed-input nonharmonic error; bubble area is controlled bidirectional leakage. (b) Mean rank over the six tasks with nontrivial harmonic support; vertical segments span each model's best-to-worst task rank. Geometry-driven implicit-topology processors tend to rank better on kernel identity, whereas TNO ranks first in mixed-input nonharmonic accuracy in all six tasks.}
\label{fig:topology-audit}
\end{figure}

Figure~\ref{fig:topology-audit} shows that these capabilities are not ordered by a single notion of topology awareness.
MGN, RIGNO, and GNOT occupy the lower-$I_H$, higher-$E_\perp^{\mathrm{mix}}$ region ($I_H=0.187/0.165/0.213$ and $E_\perp^{\mathrm{mix}}=0.848/0.894/1.040$), whereas TNO exhibits the converse profile ($I_H=0.349$, $E_\perp^{\mathrm{mix}}=0.256$).
Because GNOT uses coordinate attention rather than graph-edge message passing, this separation cannot be attributed to locality alone.
The task-wise ranks in Figure~\ref{fig:topology-audit}(b) confirm the asymmetry: MGN and RIGNO have the best average kernel-identity ranks, while TNO ranks first in mixed-input nonharmonic accuracy on all six tasks.
This profile is consistent with a generic near-identity bias in geometry-driven processors, whereas explicit incidence and harmonic coordinates may organize the representation used to predict decaying complementary dynamics.

Read together, the two panels sharpen what the explicit topology prior contributes.
TNO does not obtain its mixed-input advantage by reproducing harmonic probes more accurately: its $I_H$ is higher than those of MGN, RIGNO, and GNOT, yet it ranks first in nonharmonic component accuracy on every task.
The association of explicit harmonic information with lower complementary-component error therefore extends beyond direct kernel reconstruction, although this mixed-input metric does not isolate a pure positive-spectrum filter.
This runs counter to the narrow spectral intuition that a zero-frequency prior should help mainly with global, lowest-frequency content, while increasingly oscillatory positive-eigenvalue modes are more local and decay rapidly under diffusion.
The task-level evidence supports this broader role: positive-spectrum load $\Pi_+$ correlates with absolute OOD difficulty ($\rho_s=0.909$), and its shift correlates with overall/nonharmonic degradation ($0.608/0.800$).
Moreover, kernel-shift pressure is only weakly related to harmonic degradation ($0.148$) but more strongly related to nonharmonic degradation ($0.504$).
Thus, representing the kernel/complement split explicitly may organize predictions in the complementary component rather than merely improve the identity action on the kernel, even though the exact operator is block diagonal.

This capability asymmetry suggests a factorized design target rather than a single scalar notion of topology awareness.
A kernel branch could be constrained to identity, retaining the low $I_H$ exhibited by the geometry-driven processors, while a separate complementary-spectrum branch uses explicit incidence and harmonic coordinates to pursue TNO's low $E_\perp^{\mathrm{mix}}$; projector-based routing would control leakage between them.
Because the compared architectures differ along several dimensions, the present evidence does not identify which ingredient causes the gain, but it makes this hybrid design hypothesis directly testable.

\section{Beyond Arbitrary Geometry: Discussion, Limitations, and Outlook}

The central conclusion is that topology generalization is a distinct problem hidden by broad claims of arbitrary-geometry compatibility.
A topology shift is not merely a more extreme shape perturbation: it changes global connectivity while also altering boundaries, discretization, and the spectrum through which a physical field evolves.
In our controlled setting, the resulting difficulty is not concentrated in the invariant, lowest-frequency subspace.
The dominant error signal comes from decaying spectral content, and the model supplied with global harmonic information is strongest on that content even though it is not the most accurate at reproducing harmonic identity.
This suggests a broader multiscale interpretation: low-frequency structural information can provide context for learning higher-frequency or more rapidly decaying field components, rather than helping only the modes in which that prior is expressed directly.

The implication is not a prescription for a Hodge-specific neural architecture or evaluation protocol.
Real physical fields couple global structure and local variation in problem-dependent ways, and topology can influence long-range transport paths, recirculation, stress transmission, wave propagation, and boundary interactions.
For learning across both geometry and topology, the useful principle is therefore to connect global structural context with the local and multiscale dynamics of the field.
Such context may be represented through spectral, graph, geometric, or physics-informed features; our results do not establish that a harmonic basis, projector decomposition, or any single implementation is necessary.
They instead identify a hypothesis that can be tested more broadly: an informative low-frequency prior may improve the approximation of rapidly varying or rapidly attenuated components under topological change.

Hodge heat is well suited to exposing this hypothesis because topology is tied analytically to its null space and the remaining spectrum has an exact decay law.
That clarity is also the main limitation of the evidence.
The equation is linear, autonomous, and evaluated at a fixed time, whereas realistic systems are often nonlinear, coupled, history dependent, or driven far from equilibrium.
Homogeneous absolute boundary conditions make the cohomological structure transparent but are uncommon as a direct description of physical boundaries, which more often involve Dirichlet, Neumann, Robin, flux, interface, or mixed conditions.
TopoBox-3D further restricts geometry to box-minus-void domains, and Betti numbers omit knotting, torsion, and other distinctions.
The benchmark should therefore be read as a topology diagnostic task, not as a surrogate for performance on general physical PDEs.

Additional qualifications concern attribution and statistical scope.
The topology protocols necessarily co-vary boundary count, mesh size, and spectrum, so the matched ratios are stress-test contrasts rather than causal Betti effects.
The six processors differ in adapters and inductive biases; TNO additionally receives a per-domain nullspace basis; and equal parameter and epoch budgets do not imply equally optimized architectures.
Three seeds support median and rank-level conclusions but not fine distinctions between neighboring models.
Accordingly, the architecture comparisons and pressure correlations are associative evidence, with further boundary and statistical limitations documented in Appendix~\ref{app:limitations}.

The next step is to test the same geometry-to-topology transition on realistic physical fields, varied boundary conditions, nonlinear evolution, and time-dependent or multiphysics problems.
The central question is whether global, low-frequency information about a domain can consistently improve the learning of local, high-frequency, or rapidly decaying behavior when topology leaves the training support.
If so, topology awareness will matter not as a specialized treatment of harmonic modes, but as a way to couple global structure to multiscale physical dynamics.

\subsection*{Acknowledgements}
The authors gratefully acknowledge the financial support provided by the Beijing Nova Program (Grant No.~202604841290) and the National Natural Science Foundation of China (Grant No.~52408188).

\subsection*{AI use statement}
Generative AI tools were used to assist language drafting, code development, and review.
All mathematical statements, numerical claims, citations, and plotting code were checked against the underlying derivations, primary literature, source data, and executable analysis by the authors.
The authors take responsibility for the complete content of the paper.

\subsection*{Reproducibility statement}
Appendix~\ref{app:dataset} specifies dataset generation and audits; Appendix~\ref{app:implementation} documents model adapters, parameter counts, training, and complete results; and Appendix~\ref{app:statistics} reports resampling and diagnostic analyses.
The code repository, including the machine-readable supporting tables and deterministic scripts used to produce all reported figures, is publicly available at \url{https://github.com/asmld/TopoBox-3D}.
The TopoBox-3D dataset is publicly available at \url{https://huggingface.co/datasets/cppyyy/TopoBox-3D}.

\clearpage
\bibliography{references}
\bibliographystyle{iclr2027_conference}

\clearpage
\appendix
\section{Mathematical Details}
\label{app:math}

\subsection{Discrete Hodge decomposition and heat semigroup}

Let $B_k\in\mathbb R^{n_{k-1}\times n_k}$ denote the oriented boundary matrix from $k$-chains to $(k-1)$-chains, so that $B_kB_{k+1}=0$, and let $M_k$ be the symmetric positive-definite mass matrix on $k$-cochains (diagonal in our implementation).
With nonexistent endpoint terms omitted, the symmetric weak Hodge stiffness and the corresponding strong cochain operator are
\begin{align}
K_k
&=B_{k+1}M_{k+1}B_{k+1}^\top
  +M_kB_k^\top M_{k-1}^{-1}B_kM_k,\\
\Delta_k
&=M_k^{-1}K_k
=M_k^{-1}B_{k+1}M_{k+1}B_{k+1}^\top
  +B_k^\top M_{k-1}^{-1}B_kM_k.
\label{eq:discrete-hodge-laplacian}
\end{align}
The identities $K_k=K_k^\top\succeq0$ and
$\langle v,\Delta_kw\rangle_{M_k}=v^\top K_kw$ make $\Delta_k$ self-adjoint and positive semidefinite in the mass inner product
$\langle v,w\rangle_{M_k}=v^\top M_kw$, with norm
$\|v\|_{M_k}=(v^\top M_kv)^{1/2}$.
The compatible $k$-cochain space therefore admits the $M_k$-orthogonal decomposition
\begin{equation}
C^k(K)=\operatorname{im}B_k^\top\oplus_{M_k}\operatorname{im}(M_k^{-1}B_{k+1}M_{k+1})\oplus_{M_k}\mathcal H^k,
\end{equation}
into exact, coexact, and harmonic components.
Under homogeneous absolute boundary conditions, $\mathcal H^k=\ker\Delta_k$ and $\dim\mathcal H^k=\beta_k$.
Let $H_k$ contain an $M_k$-orthonormal basis of $\mathcal H^k$, so that
$H_k^\top M_kH_k=I$, and define
\begin{equation}
P_H=H_kH_k^\top M_k,
\qquad P_\perp=I-P_H,
\qquad
\Delta_{k,\perp}=\left.\Delta_k\right|_{(\mathcal H^k)^{\perp_{M_k}}}.
\label{eq:harmonic-projectors}
\end{equation}
Here $\Delta_{k,\perp}$ denotes the restriction to the positive spectral subspace, not the Moore--Penrose pseudoinverse.
Writing an $M_k$-orthonormal eigendecomposition with positive eigenpairs $(\lambda_i,u_i)$ gives
\begin{equation}
e^{-\kappa T\Delta_k}=P_H+\sum_{\lambda_i>0}e^{-\kappa T\lambda_i}u_iu_i^\top M_k
=P_H+e^{-\kappa T\Delta_{k,\perp}}P_\perp.
\end{equation}
Hence an exact heat operator is the identity on $\mathcal H^k$ and has no cross terms between $\mathcal H^k$ and its orthogonal complement.

For a learned map $\widehat S_T$, take controlled probes satisfying
$w_H=P_Hw_H$ and $w_\perp=P_\perp w_\perp$.
The controlled audit uses the mass-normalized probe errors
\begin{align}
I_H&=\frac{\|\widehat S_Tw_H-w_H\|_{M_k}}{\|w_H\|_{M_k}}, &
L_{H\to\perp}&=\frac{\|P_\perp\widehat S_Tw_H\|_{M_k}}{\|w_H\|_{M_k}},\\
L_{\perp\to H}&=\frac{\|P_H\widehat S_Tw_\perp\|_{M_k}}{\|w_\perp\|_{M_k}}.
\end{align}
These three quantities vanish for an exact Hodge heat operator and are evaluated on the specified finite seed-0 probe sets.
For each conventional mixed input $w_0$, we separately define the nonharmonic component error
\begin{equation}
E_\perp^{\mathrm{mix}}(w_0)=
\frac{\|P_\perp(\widehat S_Tw_0-S_Tw_0)\|_{M_k}}
{\|P_\perp S_Tw_0\|_{M_k}}.
\end{equation}
This quantity is evaluated on the standard mixed-input test records and is not the response to a pure non-harmonic probe; it can reflect both within-component approximation and input-dependent cross-component interaction.
All four quantities are summarized by medians over their corresponding records and are empirical statistics rather than induced operator norms or suprema over the full cochain space.

\subsection{Reference-domain interpretation and the matched ratio}

Let $\Phi_\theta:\Omega_0\to\Omega_\theta$ be a diffeomorphism within one fixed-topology family and $U_\theta$ the pullback on differential forms.
Then $\widetilde\Delta_{k,\theta}=U_\theta\Delta_{k,\theta}U_\theta^{-1}$ acts on the common continuous reference space over $\Omega_0$, with geometry encoded in the pulled-back metric and coefficient terms.
Compatible discretizations inherit this reference-domain interpretation through their metric-dependent mass and stiffness operators, but independently generated meshes need not have identical cochain spaces or dimensions.
Because a diffeomorphism induces an isomorphism on de Rham cohomology, the dimension of the harmonic kernel is constant within such a family.
Domains with different Betti vectors cannot be diffeomorphic, so no pullback induced by a domain diffeomorphism can identify their Hodge operators while preserving cohomology.
This obstruction motivates $Q_{m,p,k}=R_{m,p,k}/R_{m,A,k}$ as a paired comparison against fixed-topology geometry extrapolation.
This argument does not imply that equal Betti vectors guarantee diffeomorphism.

\subsection{Time discretization and Jensen spectral broadening}

The semidiscrete heat equation is
\begin{equation}
M_k\dot w(t)+\kappa K_kw(t)=0.
\end{equation}
For $N$ Crank--Nicolson steps with $\Delta t=T/N$, the generated targets satisfy
\begin{equation}
\left(M_k+\frac{\kappa\Delta t}{2}K_k\right)w^{n+1}
=\left(M_k-\frac{\kappa\Delta t}{2}K_k\right)w^n,
\qquad n=0,\ldots,N-1.
\label{eq:crank-nicolson}
\end{equation}
On a generalized eigenmode $K_ku_i=\lambda_iM_ku_i$, its amplification factor is
\begin{equation}
g_{\Delta t}(\lambda_i)
=\frac{1-\kappa\Delta t\lambda_i/2}
       {1+\kappa\Delta t\lambda_i/2}.
\end{equation}
Thus the scheme preserves harmonic modes exactly and is mass-contractive because
$|g_{\Delta t}(\lambda_i)|\leq1$ for $\lambda_i\geq0$.
For resolved modes with $\kappa\Delta t\lambda_i<2$,
\begin{equation}
g_{\Delta t}(\lambda_i)^N
=\exp\!\left[-\kappa T\lambda_i
-\frac{\kappa^3T\Delta t^2\lambda_i^3}{12}
+\mathcal O(\kappa^5T\Delta t^4\lambda_i^5)\right],
\end{equation}
which makes explicit the second-order temporal approximation to the exact heat semigroup.

For the exact semigroup, expand $w_0=\sum_i a_iu_i$ in a full $M_k$-orthonormal generalized eigenbasis over the nonnegative spectrum, including harmonic modes with $\lambda_i=0$, and set
$p_i=a_i^2/\sum_j a_j^2$.
Then
\begin{equation}
q_{T,\exp}^2
:=\frac{\|e^{-\kappa T\Delta_k}w_0\|_{M_k}^2}{\|w_0\|_{M_k}^2}
=\sum_i p_i e^{-2\kappa T\lambda_i},
\qquad
\mathcal R(w_0)=\frac{w_0^\top K_kw_0}{w_0^\top M_kw_0}
=\sum_i p_i\lambda_i.
\end{equation}
Convexity of the exponential yields the exact Jensen gap
\begin{equation}
J_{T,\exp}
:=\log q_{T,\exp}+\kappa T\mathcal R(w_0)\geq0,
\label{eq:exact-jensen-gap}
\end{equation}
with equality if and only if the spectral energy distribution is supported on a single eigenvalue.
Writing $\varkappa_r(\lambda)$ for the $r$th cumulant under $p$ gives
\begin{equation}
J_{T,\exp}
=(\kappa T)^2\operatorname{Var}_p(\lambda)
-\frac{2}{3}(\kappa T)^3\varkappa_3(\lambda)
+\mathcal O((\kappa T)^4).
\end{equation}
Hence the leading nonzero term measures spectral variance across the full nonnegative spectrum; harmonic mass at $\lambda=0$ is included rather than discarded.
The reported statistic substitutes the 100-step Crank--Nicolson retention $q_T$ for $q_{T,\exp}$ and is therefore a second-order temporal approximation to~\eqref{eq:exact-jensen-gap}.
As a numerical quality-control check, the minimum observed value over all benchmark targets is $4.33\times10^{-4}$, so the discretization does not change the nonnegativity conclusion on the analyzed samples.

\subsection{Boundary conditions}

In the convention used here, the continuous absolute boundary conditions for a sufficiently regular $k$-form $w$ are
\begin{equation}
\operatorname{tr}(\star w)=0,
\qquad
\operatorname{tr}(\star dw)=0
\qquad\text{on }\partial\Omega.
\label{eq:absolute-boundary}
\end{equation}
For $k=0$, these reduce to the homogeneous Neumann condition and leave one constant harmonic mode on each connected component.
The compatible weak discretization retains the full boundary cochain spaces and imposes these conditions naturally through the adjoint terms in~\eqref{eq:discrete-hodge-laplacian}; no essential boundary degrees of freedom are eliminated.
This choice yields harmonic spaces representing absolute de Rham cohomology and is the convention under which $\dim\mathcal H^k=\beta_k$ is used throughout the benchmark.
Relative or mixed boundary complexes admit analogous decompositions \citep{arnold2010feec,goldshtein2011mixed}, but their cohomology, harmonic representatives, and boundary traces differ.
Neither the numerical rankings nor the learned-model audit reported here has been validated under those alternatives.

\section{Dataset Details and Visualizations}
\label{app:dataset}

\subsection{Generation, splitting, and quality control}

The outer box is $[0,2]\times[0,1]^2$.
Standard tunnels are $z$-aligned cylinders with radii $[0.10,0.135]$; standard cavities are spheres with radii $[0.12,0.165]$.
In geometry family B, the horizontal displacement between the lower and upper tunnel endpoints is sampled in
$[-0.28,0.28]\times[-0.18,0.18]$, clipped to preserve the wall clearance, and constrained to have magnitude at least $0.07$.
Its ellipsoidal cavities have one semiaxis sampled from each of
$[0.12,0.14]$, $[0.16,0.19]$, and $[0.21,0.25]$; these axes are randomly permuted and the ellipsoid is rotated about a uniformly distributed direction by an angle in $[0.25,1.15]$ radians.
All other protocols use geometry family A.
Generation enforces a minimum nominal clearance of $0.10$ between voids and between cavities and the exterior, mutual disjointness, and connected material.
For tunnels, the two intersections with the $z$-faces are intentional openings, while the same clearance is enforced against the four lateral walls.

The crowded $(\beta_1,\beta_2)=(3,3)$ Test-OOD case in Protocol D uses a stratified construction rather than unconstrained rejection sampling.
Three radius-$0.10$ tunnels are initialized at horizontal coordinates
$(0.25,0.20)$, $(0.65,0.50)$, and $(1.05,0.20)$, and three radius-$0.12$ cavities at
$(1.45,0.25,0.25)$, $(1.70,0.70,0.50)$, and $(1.45,0.25,0.75)$.
The layout is randomly reflected about $y=0.5$, and each cavity center receives an independent perturbation in $[-0.012,0.012]^3$ before the same clearance and topology checks are applied.
This construction avoids rejection-sampling bias toward rare accidental packings while retaining geometric variation.

Each accepted CAD domain is meshed with linear tetrahedra using the Gmsh HXT three-dimensional algorithm.
The target element size is $0.070$ near tunnel and cavity boundaries, increases to $0.115$ in the bulk over a refinement distance of $0.22$, and is followed by Netgen mesh optimization.
Tunnel cylinders are extended by $0.15$ beyond each intended opening before Boolean subtraction, preventing artificial end caps and small intersection slivers for tilted tunnels.
The resulting oriented tetrahedral complex uses canonically ordered edges and faces and positively oriented tetrahedra.
Each accepted geometry is tetrahedralized and audited for Betti numbers, Euler characteristic, nullities, $B_1B_2=0$, $B_2B_3=0$, finite fields, energy fractions, and heat-contractivity constraints.
The formal audit reports zero errors over 5,280 geometries and 63,360 PDE instances.
Within every training, validation, and Test-IID split, the admissible Betti pairs in Table~\ref{tab:protocols} are allocated as evenly as the split size permits, so their counts differ by at most one.

\begin{table}[H]
\centering
\caption{Geometry counts per protocol. All fields and initial-condition regimes associated with one geometry remain in the same split.}
\begin{tabular}{@{}lrrrrr@{}}
\toprule
Split & A & B & C & D & Total \\
\midrule
Train & 800 & 800 & 800 & 800 & 3,200 \\
Validation & 120 & 120 & 120 & 120 & 480 \\
Test-IID & 200 & 200 & 200 & 200 & 800 \\
Test-OOD & 200 & 200 & 200 & 200 & 800 \\
\midrule
Total & 1,320 & 1,320 & 1,320 & 1,320 & 5,280 \\
\bottomrule
\end{tabular}
\end{table}

\begin{figure}[H]
\centering
\includegraphics[width=\linewidth]{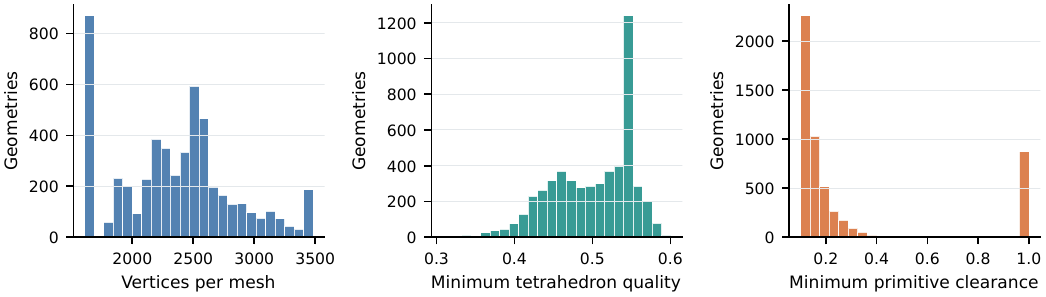}
\caption{Distributions computed from the formal 5,280-geometry manifest: mesh size, minimum tetrahedron quality, and primitive clearance. The mass at clearance 1 corresponds to geometries for which the relevant pairwise primitive-clearance constraint is vacuous.}
\label{fig:dataset-qc}
\end{figure}

\FloatBarrier
\subsection{Fields and initial conditions}

For $k=0,1,2$, tokens are vertices, oriented edges, and oriented faces.
Vertex coordinates are normalized by the box side lengths.
The shared vertex geometry channels are normalized $(x,y,z)$, a binary boundary indicator, and an analytic signed-distance field that is positive in the material and negative inside a void or outside the box.
For $k=1$ and $k=2$, token coordinates are simplex barycenters; the five vertex channels are averaged over the incident vertices and concatenated with the oriented edge vector or oriented face-area vector and with edge length or face area.
Edges and faces use increasing vertex order as the canonical orientation.
Including the active scalar cochain value gives six input channels for $k=0$ and ten for $k=1,2$.
Generic baselines do not receive Betti numbers, cross-degree incidence, or harmonic bases.
Only TNO receives all degree-wise cochains, $B_1,B_2,B_3$, and the active $M_k$-orthonormal harmonic basis.

Initial conditions are generated from smooth ambient Gaussian radial-basis fields with eight centers and correlation length $0.40$.
For $k=1,2$, a smooth $(k-1)$-cochain potential $\phi$ and a smooth $(k+1)$-cochain potential $\psi$ produce
\begin{equation}
w_E=B_k^\top\phi,
\qquad
w_C=M_k^{-1}B_{k+1}M_{k+1}\psi.
\label{eq:initial-hodge-components}
\end{equation}
Mesh-scale content is suppressed by applying three passes of the subspace-preserving resolvent filter
\begin{equation}
w\leftarrow(M_k+0.10K_k)^{-1}M_kw,
\label{eq:initial-resolvent-filter}
\end{equation}
followed by mass-orthogonal cleanup and unit-$M_k$ normalization of each realizable component.
When $\mathcal H^k$ is nontrivial, an independent Gaussian coefficient vector in the $M_k$-orthonormal harmonic basis is normalized to produce $w_H$.
For requested energy fractions $(\rho_E,\rho_C,\rho_H)$, the final unit-mass initial condition is
\begin{equation}
w_0=\sqrt{\rho_E}\,w_E+\sqrt{\rho_C}\,w_C+\sqrt{\rho_H}\,w_H.
\end{equation}

\begin{table}[H]
\centering
\caption{Requested exact/coexact/harmonic energy fractions for $k=1,2$.}
\begin{tabular}{@{}lccc@{}}
\toprule
Configuration & $\rho_E$ & $\rho_C$ & $\rho_H$ \\
\midrule
Non-harmonic & $1/2$ & $1/2$ & $0$ \\
Weak harmonic & $4/9$ & $4/9$ & $1/9$ \\
Balanced & $1/3$ & $1/3$ & $1/3$ \\
Strong harmonic & $1/4$ & $1/4$ & $1/2$ \\
\bottomrule
\end{tabular}
\end{table}

These are requested rather than universally realized fractions.
If $\dim\mathcal H^k=0$, the unavailable harmonic energy is redistributed equally and all four configurations realize $(\rho_E,\rho_C,\rho_H)=(1/2,1/2,0)$.
For $k=0$, the four configurations instead use independent smooth scalar fields whose mass-weighted mean is removed; the constant harmonic mode is suppressed and the realized split is $(0,1,0)$.

Raw face-cochain coefficients are smaller in magnitude because they represent integrals over oriented face areas.
For $k=2$ only, every adapter therefore supplies the model with $w_0/s$ where
\begin{equation}
s=\max\!\left\{\left(\frac{1}{n_2}\sum_{j=1}^{n_2}w_{0,j}^2\right)^{1/2},10^{-8}\right\}.
\label{eq:k2-cochain-scaling}
\end{equation}
The native prediction is multiplied by the same $s$ before computing the mass-weighted loss and all reported errors; $k=0,1$ use $s=1$.
Thus this conditioning step does not change the physical cochain units or the relative-error definition.

\FloatBarrier
\section{Implementation and Complete Model Results}
\label{app:implementation}
\raggedbottom

\subsection{Adapters, optimization, and completeness}

MGN constructs same-degree adjacency by shared cofaces.
RIGNO retains its official region-interaction processor and constructs sample-dependent physical--regional graphs.
Transolver uses 32 physics slices and six attention layers.
GNOT uses coordinate-aligned trunk tokens, one branch function, and no graph-edge computation in the processor.
GAOT uses its MAGNO encoder/decoder and a 128-token latent grid; for oriented cochains, a local orientation-aware residual decoder is added to the coordinate-only native decoding path.
TNO is a project implementation of the published cross-degree incidence and harmonic mechanism, not author code.

Table~\ref{tab:exact-model-configs} gives the instantiated capacity-matching configurations.
Parameter counts include every trainable input adapter and output readout; the small $k=0$ versus $k=1,2$ differences arise from the degree-dependent input channels in Appendix~\ref{app:dataset}.
TNO instantiates all three degree streams jointly and therefore has the same count for every active degree.

\begin{table}[H]
\centering
\scriptsize
\caption{Exact model configurations and trainable parameter counts. The $k=1$ and $k=2$ counts coincide for every model.}
\label{tab:exact-model-configs}
\begin{tabular}{@{}lp{0.49\linewidth}rr@{}}
\toprule
Model & Core configuration & $k=0$ & $k=1,2$ \\
\midrule
MGN & width 160; 6 message-passing layers & 1,162,241 & 1,162,881 \\
RIGNO & node/edge width 128; 7 processor steps & 1,197,059 & 1,198,083 \\
Transolver & width 192; 6 layers; 8 heads; 32 slices & 1,210,033 & 1,211,569 \\
GNOT & width 96; 4 layers; 8 heads; 2 experts & 1,134,921 & 1,135,689 \\
GAOT & MAGNO width 48; transformer width 128 and 4 layers; $8\!\times\!4\!\times\!4$ latent grid & 1,185,645 & 1,186,102 \\
TNO & width 170; 4 cross-degree layers; harmonic branch active & 1,167,563 & 1,167,563 \\
\bottomrule
\end{tabular}
\end{table}

All 216 runs reached 300 epochs.
Validation is evaluated every five epochs, early stopping is disabled, and the best validation checkpoint is used for final IID/OOD evaluation.
PyTorch models use the same cosine schedule; RIGNO uses the corresponding Optax schedule over the full update count.
MGN, Transolver, GNOT, and GAOT use BF16 autocast, TNO uses FP32 sparse operations, and RIGNO follows the official JAX numerical path.
Because meshes have variable simplex counts, every optimizer step contains one geometry.
During training, each epoch visits all 800 geometries once and uses one of the four initial-condition configurations per geometry; the configuration index cycles deterministically with epoch, geometry, and seed, so each geometry is paired with all four configurations in every consecutive four-epoch block.
Validation and both test splits evaluate all four configurations for every geometry.

\FloatBarrier
\subsection{Complete degradation ratios}

Tables~\ref{tab:iid-loss} and~\ref{tab:ood-loss} report the complete median Test-IID and Test-OOD losses in the main paper.
Table~\ref{tab:degradation-ratio} gives the corresponding degradation ratio $R$, formed within each seed before taking the three-seed median.
Thus each entry is $\operatorname{median}_{s}\!\bigl(E^{\mathrm{OOD}}_{s}/E^{\mathrm{IID}}_{s}\bigr)$ rather than the ratio of two separately pooled medians.
Values below one mean only that the sampled OOD split has lower aggregate error for that model--task cell; they do not by themselves imply a beneficial topology shift, which is why the main analysis also reports the Protocol-A-matched ratio $Q$.

\begin{table}[H]
\centering\scriptsize
\caption{OOD/IID degradation ratio $R$.}
\label{tab:degradation-ratio}
\resizebox{\linewidth}{!}{\begin{tabular}{@{}lrrrrrrrrrrrr@{}}
\toprule
& A0&A1&A2&B0&B1&B2&C0&C1&C2&D0&D1&D2\\\midrule
MGN&1.037&1.048&1.385&1.507&1.197&1.265&1.179&1.150&1.362&1.292&1.147&1.129\\
RIGNO&1.027&1.002&1.136&1.259&1.041&1.211&1.027&1.061&1.220&1.108&1.068&1.169\\
Transolver&1.066&.993&1.179&1.407&1.073&1.322&1.059&1.019&1.293&1.131&1.057&1.167\\
GNOT&1.094&1.011&1.156&1.912&1.477&1.353&1.229&1.206&1.374&1.377&1.247&1.263\\
GAOT&1.005&.980&1.008&.986&.990&1.326&.994&1.008&1.187&.988&.988&1.161\\
TNO&1.039&1.051&1.005&1.496&.878&1.351&1.206&1.156&.776&1.218&.858&.894\\
\bottomrule\end{tabular}}
\end{table}

\FloatBarrier
\subsection{Representative predictions}

Qualitative visualizations are selected deterministically from the evaluation tables: a typical example is closest to the split median error and a challenging example is closest to a high-error quantile.
They are illustrative and are not used to rank models.
More precisely, selection is restricted to the balanced initial condition and uses the mean relative $L^2$ error pooled over all six models and three seeds: the IID and OOD typical geometries are closest to the respective 50th percentiles, and the challenging OOD geometry is closest to the 85th percentile.
The displayed predictions use the selected model's seed-0 best-validation checkpoint, while the model-independent selection rule prevents choosing a geometry specifically favorable to that model.

For $k=0$, color denotes the signed vertex cochain directly.
For $k=1$ and $k=2$, the edge or face cochain is reconstructed as a vertex vector field by local ridge-regularized least squares against the oriented edge or face-area vectors and is then shown with streamlines.
This reconstruction is used only for visualization: every reported error is computed on the original oriented cochain with the mass matrix in Eq.~\eqref{eq:losses}.
Within each panel row, target and prediction share a magnitude scale truncated at their pooled 99th percentile, whereas the error has its own 99th-percentile scale; streamline color therefore supports spatial comparison but not direct target-to-error magnitude comparison.
Translucent orange and green surfaces mark tunnels and cavities, respectively.

\begin{figure}[H]
\centering
\includegraphics[width=\linewidth]{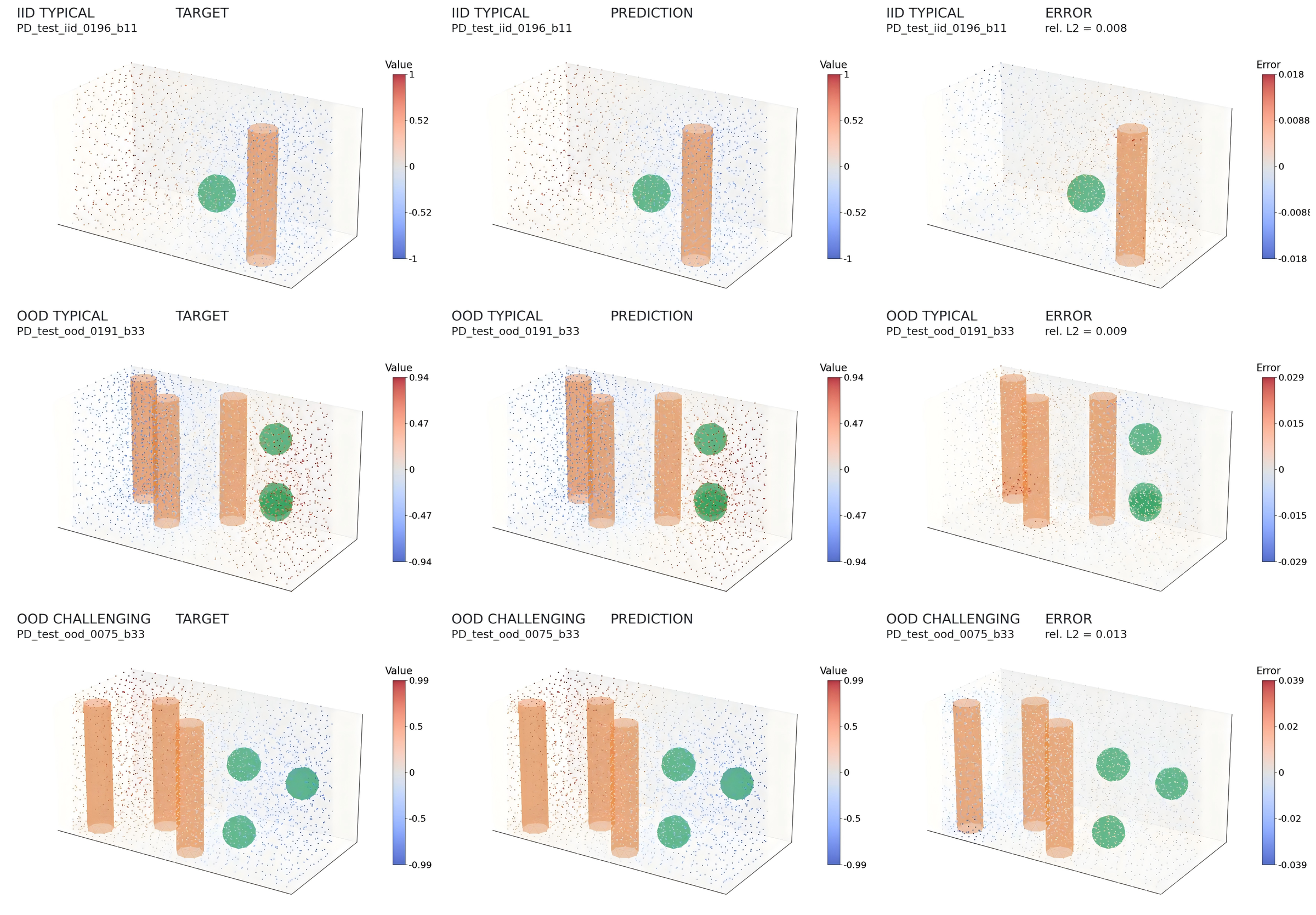}
\caption{MGN on Protocol D, $k=0$, under the balanced initial condition: target, prediction, and signed error for typical IID, typical topology-OOD, and challenging topology-OOD examples.}
\label{fig:representative-k0}
\end{figure}

\begin{figure}[p]
\centering
\includegraphics[width=\linewidth]{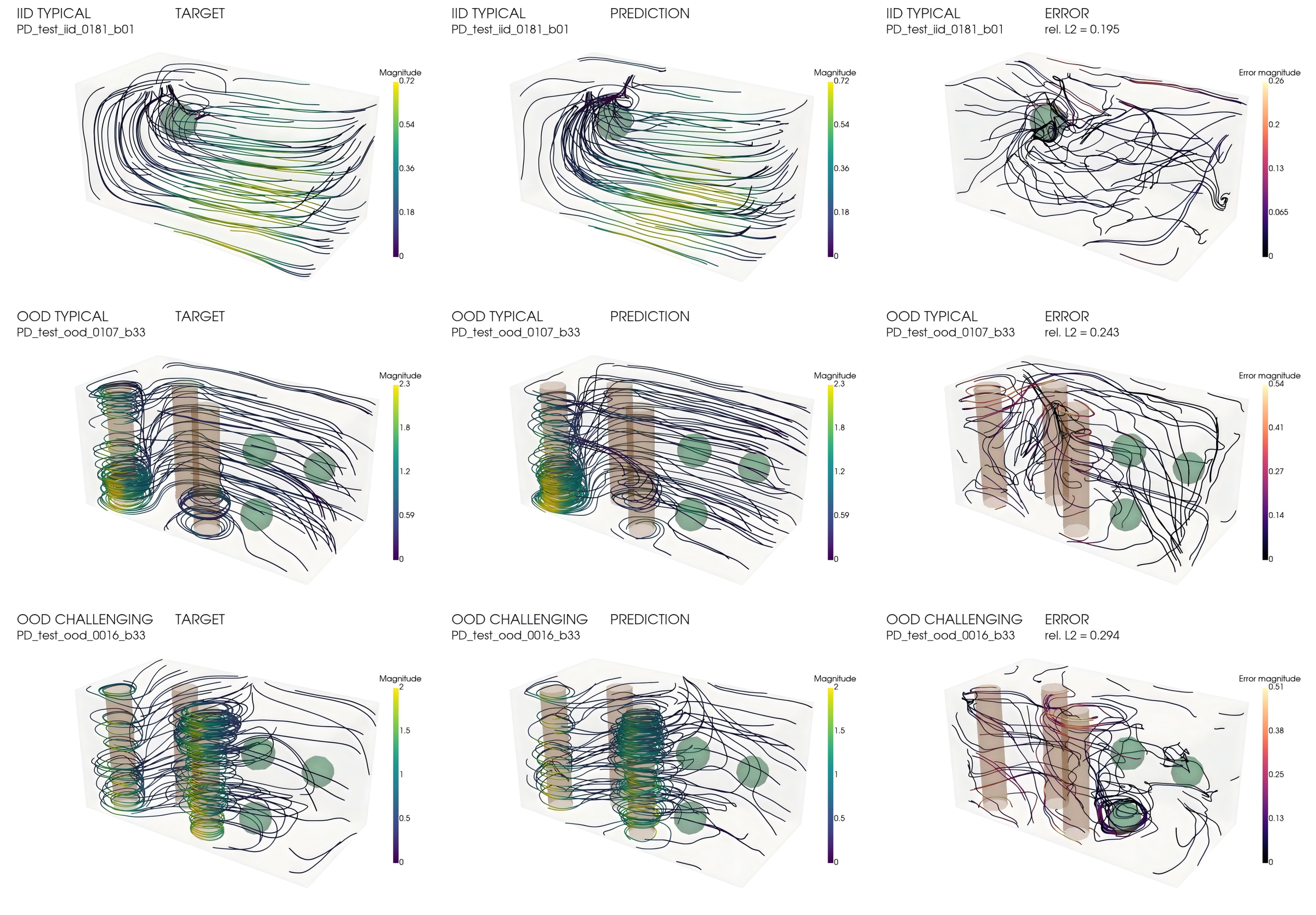}
\caption{MGN on Protocol D, $k=1$, under the balanced initial condition. Edge cochains are rendered through reconstructed vector-field streamlines using the same three-row selection convention described above.}
\label{fig:representative-k1}
\end{figure}

\begin{figure}[p]
\centering
\includegraphics[width=\linewidth]{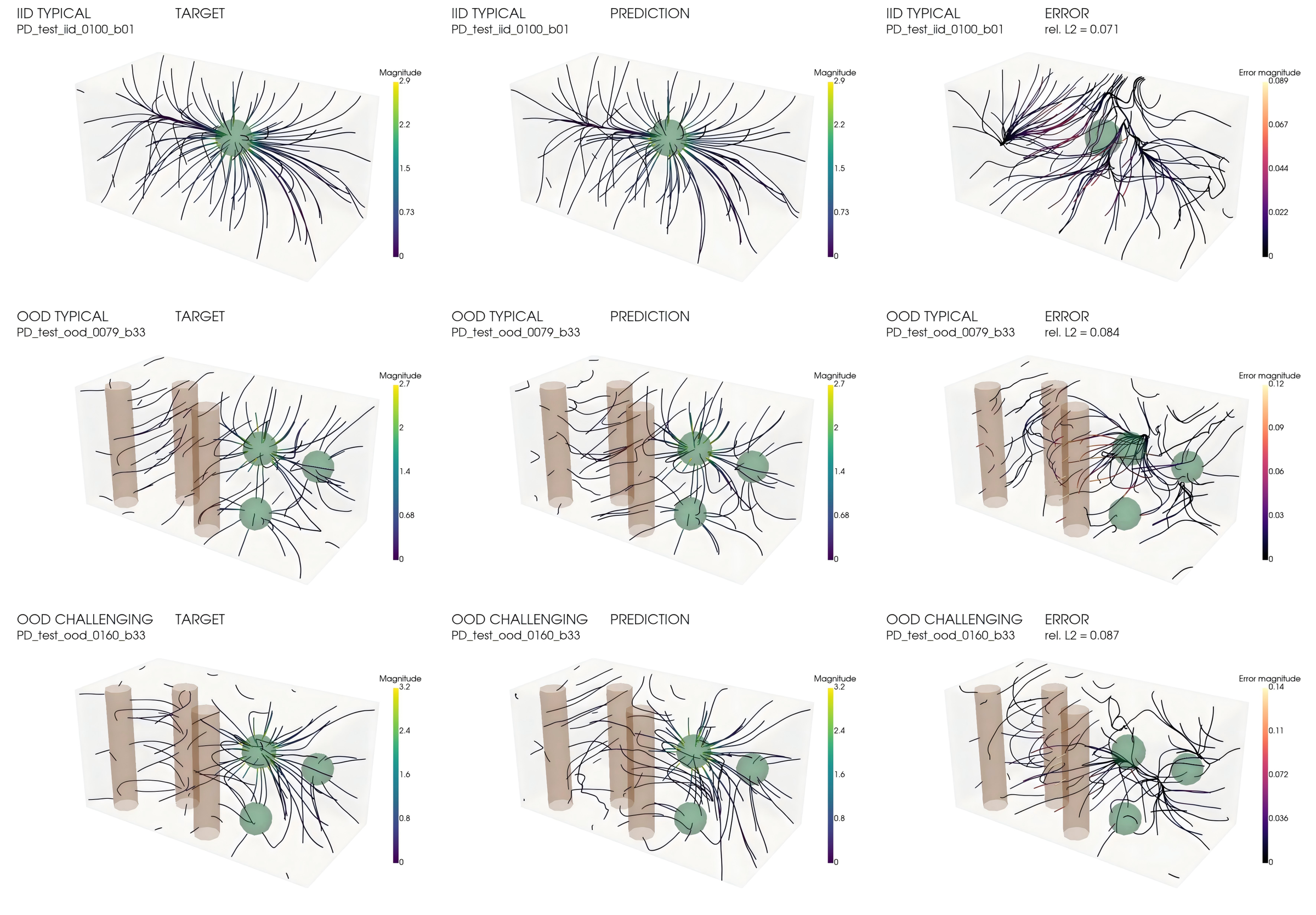}
\caption{TNO on Protocol D, $k=2$, under the balanced initial condition. Face cochains are rendered through reconstructed vector-field streamlines using the same three-row selection convention described above.}
\label{fig:representative-k2}
\end{figure}

\FloatBarrier
\flushbottom
\section{Additional Statistical Analyses and Topology Audit}
\label{app:statistics}

\subsection{Complete matched topology penalties}

Section~\ref{sec:geometry-topology} reports that excess topology-OOD degradation is widespread across protocols and degrees, rather than concentrated in the degree whose harmonic dimension changes.
For completeness, Tables~\ref{tab:complete-q-b}--\ref{tab:complete-q-d} give every matched penalty $Q_{m,p,k}=R_{m,p,k}/R_{m,A,k}$ from Figure~\ref{fig:performance-topology} together with its geometry-cluster 95\% interval.
Here $Q>1$ means that the topology-OOD protocol produces greater relative IID-to-OOD degradation than the degree-matched fixed-topology geometry shift in Protocol A, whereas $Q<1$ means less matched degradation.
These quantities are stress-test contrasts, not causal estimates of a Betti-number effect, because protocols B--D also change boundary count, discretization, and the positive spectrum.

Across the five models without the full chain complex and harmonic basis, 37/45 estimates exceed one and the geometric mean is 1.110.
The same direction occurs in 17/20 cells from B-$k=1$, C-$k=2$, and D-$k=1,2$, where $\dim\ker\Delta_k$ changes, but their geometric mean is only 1.083.
Thus, changing the active harmonic dimension does not amplify the matched penalty relative to the all-degree result.
TNO is heterogeneous across all nine B/C/D tasks (5/9 estimates above one; geometric mean 1.034), while its four harmonic-dimension-changing comparisons are all below one (geometric mean 0.827).
This model-specific reversal reinforces the broader conclusion that topology-OOD difficulty is not localized to the harmonic kernel.

\begin{table}[H]
\centering
\caption{Summary of matched topology penalties. ``Harmonic dimension changes'' comprises B-$k=1$, C-$k=2$, and D-$k=1,2$. Geometric means are computed by averaging $\log Q$.}
\label{tab:aggregate-topology-penalty}
\resizebox{\linewidth}{!}{%
\begin{tabular}{@{}lrrrrr@{}}
\toprule
Scope & Cells & $Q>1$ & CI $>1$ & CI $<1$ & Geom. mean $Q$\\\midrule
Five implicit-topology models, all B/C/D degrees &45&37&29&5&1.110\\
Five implicit-topology models, harmonic dimension changes &20&17&13&1&1.083\\
TNO, all B/C/D degrees &9&5&5&4&1.034\\
TNO, harmonic dimension changes &4&0&0&4&0.827\\
\bottomrule
\end{tabular}%
}
\end{table}

In the cell-wise tables below, $\uparrow$ and $\downarrow$ indicate 95\% geometry-cluster intervals entirely above and below one, respectively; unmarked intervals cross one.
To pair every interval with the estimator that generated it, the tabulated point is the mean $\overline Q_{\mathrm{boot}}$ over 1,000 independent geometry-cluster resamples of the 200 IID and 200 OOD geometry IDs in Protocol A and the compared protocol, retaining all four initial conditions for each selected geometry.
The summary counts and geometric means above continue to use the three-seed median $Q$ defined in Equation~\ref{eq:rq}; the two point summaries agree to the reported qualitative conclusions but are labeled separately.

\begin{table}[H]
\centering\scriptsize
\caption{Complete Protocol B matched penalties. Each cell is $\overline Q_{\mathrm{boot}}$ [95\% geometry-cluster interval].}
\label{tab:complete-q-b}
\begin{tabular}{@{}lccc@{}}
\toprule
Model & $k=0$ & $k=1$ & $k=2$\\\midrule
MGN & 1.448 [1.325, 1.592]$\uparrow$ & 1.141 [1.099, 1.183]$\uparrow$ & 0.919 [0.874, 0.968]$\downarrow$\\
RIGNO & 1.228 [1.111, 1.344]$\uparrow$ & 1.018 [0.997, 1.041] & 1.075 [1.039, 1.110]$\uparrow$\\
Transolver & 1.324 [1.194, 1.465]$\uparrow$ & 1.070 [1.049, 1.093]$\uparrow$ & 1.132 [1.086, 1.181]$\uparrow$\\
GNOT & 1.795 [1.631, 1.989]$\uparrow$ & 1.532 [1.485, 1.588]$\uparrow$ & 1.169 [1.120, 1.220]$\uparrow$\\
GAOT & 0.982 [0.971, 0.994]$\downarrow$ & 1.010 [0.997, 1.023] & 1.316 [1.266, 1.363]$\uparrow$\\
TNO & 1.438 [1.314, 1.562]$\uparrow$ & 0.833 [0.794, 0.874]$\downarrow$ & 1.344 [1.279, 1.408]$\uparrow$\\
\bottomrule
\end{tabular}
\end{table}

\begin{table}[H]
\centering\scriptsize
\caption{Complete Protocol C matched penalties. Each cell is $\overline Q_{\mathrm{boot}}$ [95\% geometry-cluster interval].}
\label{tab:complete-q-c}
\begin{tabular}{@{}lccc@{}}
\toprule
Model & $k=0$ & $k=1$ & $k=2$\\\midrule
MGN & 1.135 [1.043, 1.236]$\uparrow$ & 1.094 [1.044, 1.146]$\uparrow$ & 0.993 [0.944, 1.045]\\
RIGNO & 1.003 [0.919, 1.094] & 1.061 [1.031, 1.091]$\uparrow$ & 1.084 [1.048, 1.121]$\uparrow$\\
Transolver & 0.992 [0.895, 1.100] & 1.016 [0.992, 1.040] & 1.105 [1.059, 1.150]$\uparrow$\\
GNOT & 1.124 [1.024, 1.220]$\uparrow$ & 1.185 [1.149, 1.220]$\uparrow$ & 1.189 [1.140, 1.243]$\uparrow$\\
GAOT & 0.989 [0.978, 1.000]$\downarrow$ & 1.029 [1.012, 1.045]$\uparrow$ & 1.178 [1.138, 1.222]$\uparrow$\\
TNO & 1.161 [1.068, 1.265]$\uparrow$ & 1.099 [1.048, 1.154]$\uparrow$ & 0.771 [0.728, 0.820]$\downarrow$\\
\bottomrule
\end{tabular}
\end{table}

\begin{table}[H]
\centering\scriptsize
\caption{Complete Protocol D matched penalties. Each cell is $\overline Q_{\mathrm{boot}}$ [95\% geometry-cluster interval].}
\label{tab:complete-q-d}
\begin{tabular}{@{}lccc@{}}
\toprule
Model & $k=0$ & $k=1$ & $k=2$\\\midrule
MGN & 1.246 [1.137, 1.362]$\uparrow$ & 1.091 [1.055, 1.130]$\uparrow$ & 0.822 [0.781, 0.866]$\downarrow$\\
RIGNO & 1.083 [0.987, 1.189] & 1.046 [1.023, 1.070]$\uparrow$ & 1.022 [0.990, 1.057]\\
Transolver & 1.058 [0.958, 1.178] & 1.071 [1.051, 1.092]$\uparrow$ & 0.988 [0.949, 1.025]\\
GNOT & 1.244 [1.127, 1.366]$\uparrow$ & 1.237 [1.200, 1.279]$\uparrow$ & 1.091 [1.047, 1.139]$\uparrow$\\
GAOT & 0.983 [0.973, 0.994]$\downarrow$ & 1.006 [0.993, 1.018] & 1.153 [1.118, 1.191]$\uparrow$\\
TNO & 1.171 [1.077, 1.273]$\uparrow$ & 0.815 [0.771, 0.866]$\downarrow$ & 0.888 [0.823, 0.955]$\downarrow$\\
\bottomrule
\end{tabular}
\end{table}

\FloatBarrier
\subsection{Full driver analysis}

\begin{figure}[H]
\centering
\includegraphics[width=\linewidth]{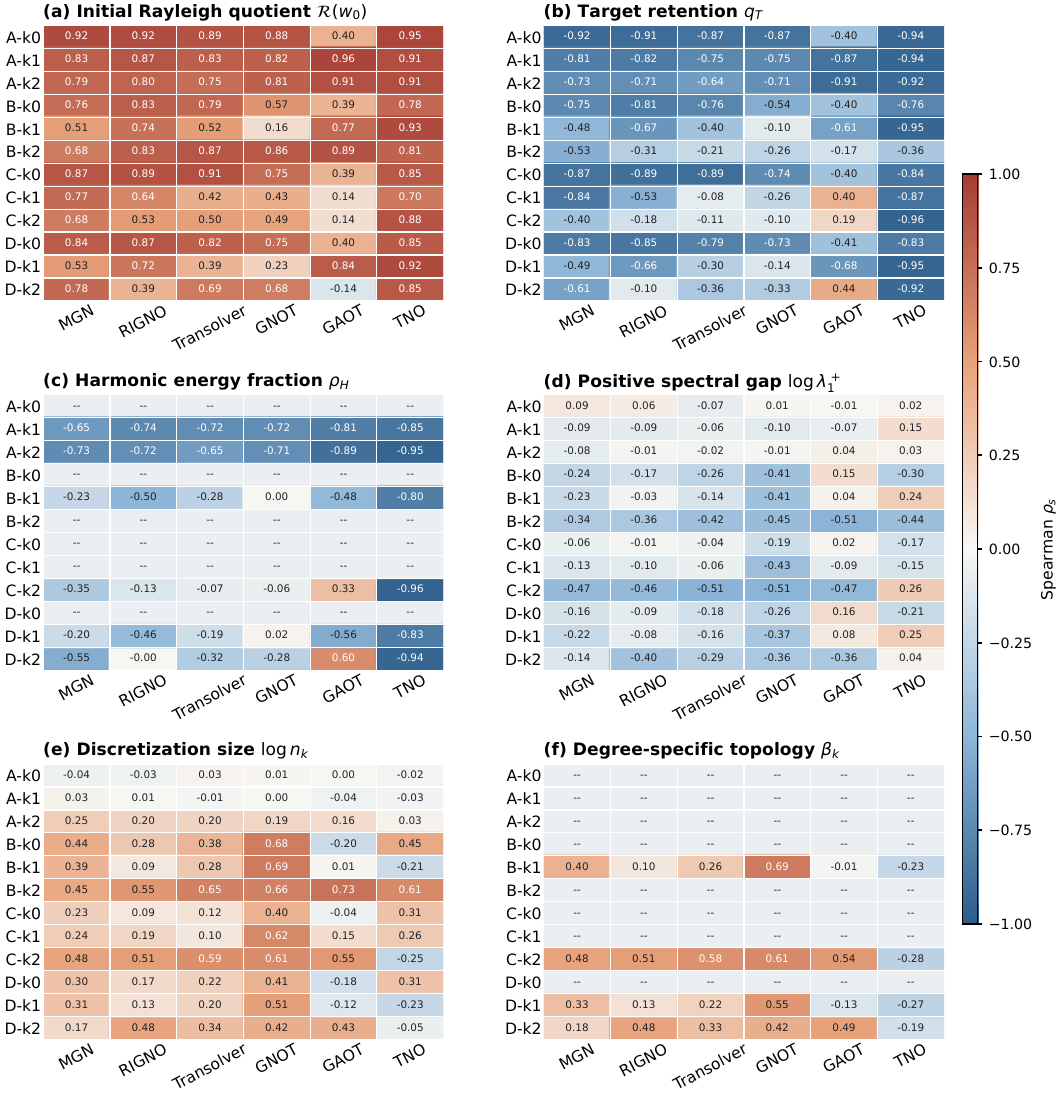}
\caption{Cell-wise Spearman correlations for six principal descriptors. Rows are protocol--degree tasks and columns are models. Dashed cells indicate constant, unavailable predictors.}
\label{fig:driver-heatmaps}
\end{figure}

Each of the 72 model--protocol--degree cells contains 1,600 test instances from 400 geometries after taking the median prediction error over seeds.
Spearman correlations, linear $R^2$, standardized slopes, and geometry-cluster standard errors are computed separately for seven candidate variables.
Constant predictors are marked unavailable rather than assigned zero association.

The initial Rayleigh quotient $\mathcal R(w_0)$ is the most consistent predictor of relative error: across the 72 cells, its median Spearman correlation is 0.784, its median single-variable $R^2$ is 0.585, and it gives the largest $R^2$ in 43/72 cells.
Target retention $q_T$ is the strongest negative predictor (median $\rho_s=-0.693$, median $R^2=0.455$), but it is not an independent mechanism because the target norm also enters the denominator of the reported relative error.
For the 36 estimable higher-order cells, harmonic energy fraction is consistently protective (median $\rho_s=-0.521$, median $R^2=0.257$; all 36 geometry-cluster tests have $p<0.05$), as expected from exact preservation of the zero-eigenvalue component.

The remaining structural descriptors are weaker globally.
The median $R^2$ values for $\log n_k$, $\beta_k$, $\beta_1+\beta_2$, and $\log\lambda_1^+$ are 0.048, 0.106, 0.067, and 0.012, respectively, but increase to 0.195, 0.189, 0.252, and 0.143 at $k=2$.
Thus, discretization size, topological complexity, and the low end of the positive spectrum become more informative for face cochains even though none is a dominant cross-task predictor.
Blank or dashed cells in Figure~\ref{fig:driver-heatmaps} denote a descriptor that is constant within that model--task cell and therefore has no estimable within-cell correlation; they do not denote zero association.

The Jensen gap $J_T=\log q_T+\kappa T\mathcal R(w_0)$ measures spectral dispersion beyond the Rayleigh mean.
Its marginal correlation with relative error is positive in most cells (median $\rho_s=0.503$), but after controlling for $\mathcal R(w_0)$ the median partial correlation is only 0.009 and the median incremental rank-$R^2$ is 0.010.
The independent contribution is degree dependent: for $k=0,1,2$, the median partial correlations are $-0.032$, 0.100, and 0.121, and the corresponding incremental rank-$R^2$ values are 0.0003, 0.038, and 0.086.
The strongest higher-order examples are C-$k=2$ and D-$k=2$, whose median partial correlations are 0.609 and 0.427 and whose median incremental rank-$R^2$ values are 0.302 and 0.130.
By contrast, B-$k=2$ has a strong marginal correlation (median 0.756) but almost no information beyond the Rayleigh quotient (median partial correlation $-0.023$, incremental rank-$R^2=0.005$).
These results identify spectral broadening as an architecture- and task-dependent higher-order correction rather than a universal error mechanism.

\begin{figure}[H]
\centering
\includegraphics[width=\linewidth]{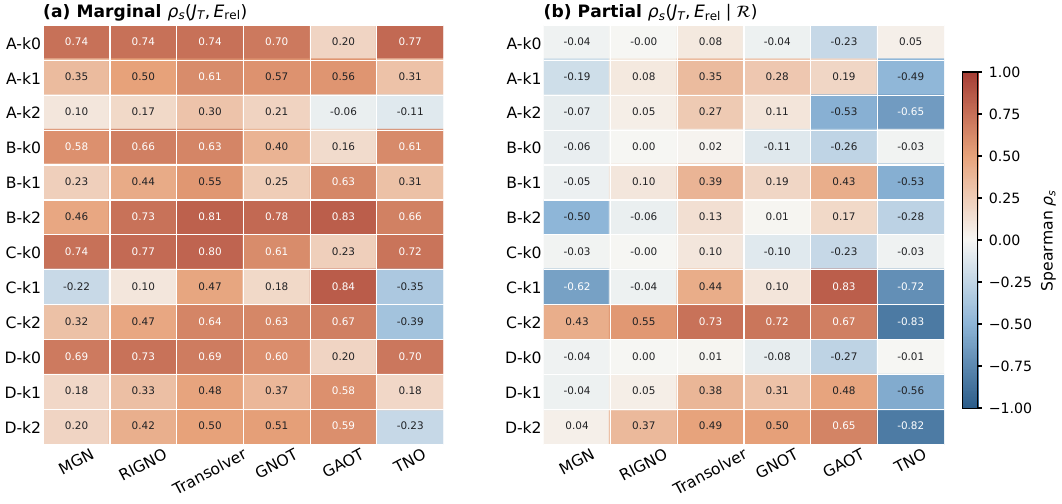}
\caption{Marginal and conditional Jensen-gap correlations by model and task. The left panel reports $\rho_s(J_T,E_{\rm rel})$; the right panel reports the partial Spearman correlation after controlling for $\mathcal R(w_0)$. Conditioning removes most of the global association but leaves task-dependent higher-order structure.}
\label{fig:jensen-heatmaps}
\end{figure}

\FloatBarrier
\subsection{Extended topology-pressure diagnostics}

We separate \emph{task-side pressure}, which describes what an OOD task demands, from \emph{model-side capability}, which describes how a trained predictor acts on the harmonic and positive spectral subspaces.
Compressing both into a single difficulty score would obscure whether an error is associated with a changing kernel, a difficult OOD spectrum, or a shift of that spectrum away from training support.

\paragraph{Task-side pressure coordinates.}
For each protocol--degree task we use the three-coordinate vector
\begin{equation}
\boldsymbol\Pi=(\Pi_H,\Pi_+,\Delta\Pi_+).
\end{equation}
The kernel-shift coordinate combines the IID--OOD Wasserstein distances of the harmonic dimension and harmonic input fraction,
\begin{equation}
\Pi_H=\left[
\left(\frac{W_1(p_{\rm IID}(\beta_k),p_{\rm OOD}(\beta_k))}{3}\right)^2+
\left(\frac{W_1(p_{\rm IID}(\rho_H),p_{\rm OOD}(\rho_H))}{s_{\rho_H}}\right)^2
\right]^{1/2},
\end{equation}
where $s_{\rho_H}$ is a robust global scale.
The positive-spectrum load $\Pi_+$ combines the across-task percentile ranks of the OOD medians of $\mathcal R(w_0)$ and $J_T$, and therefore measures absolute positive-spectrum difficulty.
The positive-spectrum shift
\begin{equation}
\Delta\Pi_+=\operatorname{RMS}\!\left(
\widetilde W_1(\mathcal R),\widetilde W_1(J_T),
\widetilde W_1(\log\lambda_1^+),\widetilde W_1(\log n_k)
\right)
\end{equation}
instead measures how far the OOD spectral/discretization distribution moves from IID support; each $\widetilde W_1$ is robustly normalized before aggregation.

\begin{figure}[!t]
\centering
\includegraphics[width=\linewidth]{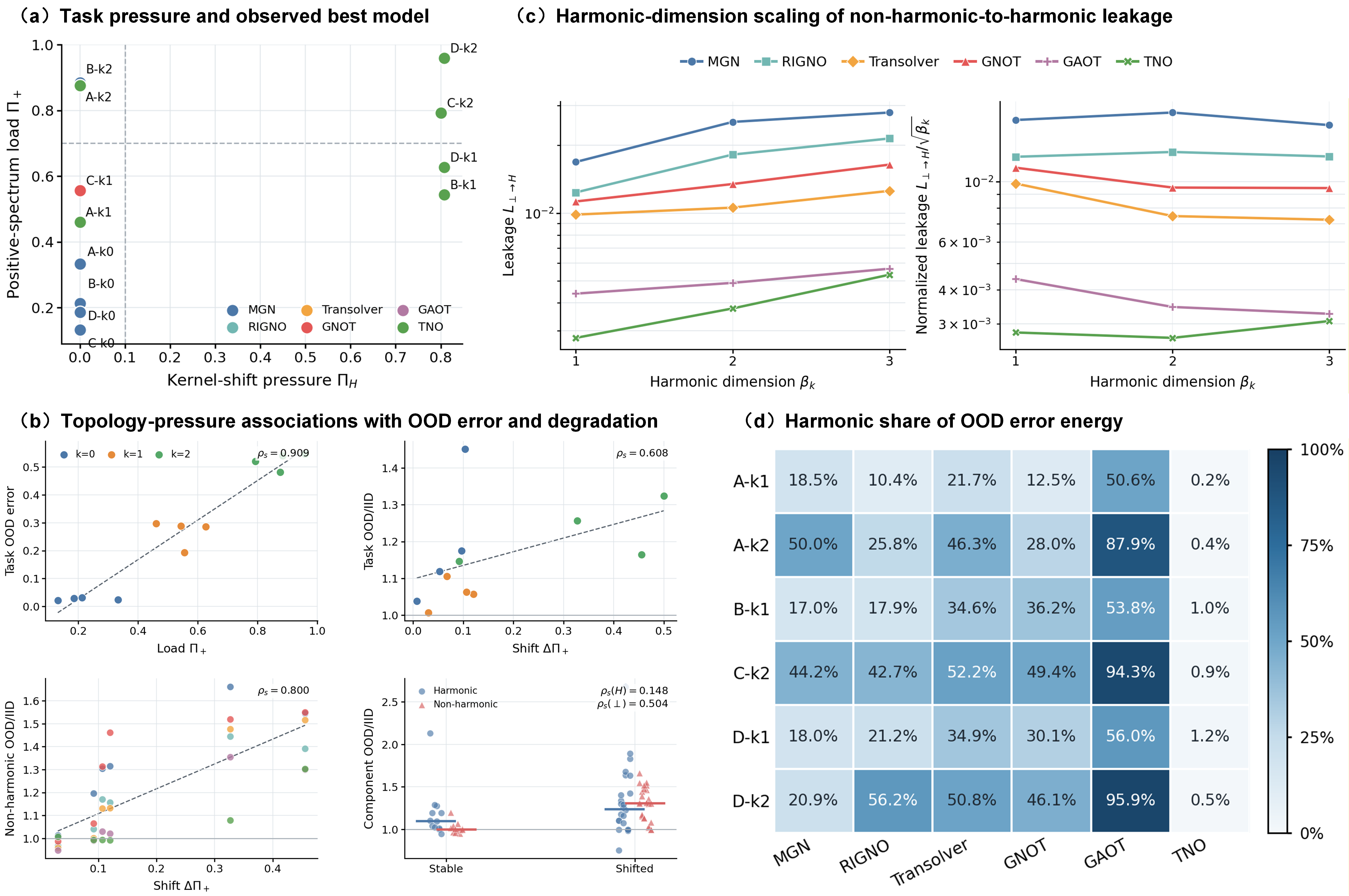}
\caption{Extended topology audit. (a) The 12 tasks in the $(\Pi_H,\Pi_+)$ pressure plane, colored by the observed best model; dashed lines mark the pressure-group thresholds used for interpretation. (b) Pressure--error associations, read left-to-right and top-to-bottom: $\Pi_+$ versus task OOD error; $\Delta\Pi_+$ versus task OOD/IID ratio; $\Delta\Pi_+$ versus non-harmonic OOD/IID ratio; and component OOD/IID ratios for kernel-stable and kernel-shifted tasks. In the final panel, circles and triangles denote harmonic and non-harmonic components, respectively, and horizontal bars show group medians; all reported $\rho_s$ values are Spearman correlations. (c) Pure non-harmonic-to-harmonic leakage: total $L_{\perp\to H}$ on the left and the dimension-normalized quantity $L_{\perp\to H}/\sqrt{\beta_k}$ on the right. (d) Percentage of conventional mixed-input OOD error energy assigned to the harmonic component for each task--model pair.}
\label{fig:extended-topology-audit}
\end{figure}

Figure~\ref{fig:extended-topology-audit}(a) places the 12 tasks in the $(\Pi_H,\Pi_+)$ plane and colors each point by its observed best model.
The four kernel-changing tasks B-$k=1$, C-$k=2$, and D-$k=1,2$ all have $\Pi_H\simeq0.8$, whereas the remaining tasks have $\Pi_H\simeq0$.
The two resulting vertical bands are therefore a property of the benchmark design, not a plotting error; $\Pi_H$ is a near-binary kernel-shift indicator here and should not be interpreted as a universal continuous topology-complexity measure.

Figure~\ref{fig:extended-topology-audit}(b) validates the pressure coordinates at two statistical levels.
Across the 12 task medians, $\Pi_+$ strongly tracks absolute OOD error ($\rho_s=0.909$, $p=4.2\times10^{-5}$), whereas $\Delta\Pi_+$ tracks the task-level OOD/IID ratio ($\rho_s=0.608$, $p=0.036$).
Across 36 model--task non-harmonic cells, $\Delta\Pi_+$ is more strongly associated with positive-branch degradation ($\rho_s=0.800$, $p=4.8\times10^{-9}$).
For the kernel comparison, the near-binary $\Pi_H$ is shown categorically rather than as a continuous scatter: its association with harmonic degradation is weak and non-significant ($\rho_s=0.148$, $p=0.389$), but its association with non-harmonic degradation is stronger ($\rho_s=0.504$, $p=0.0017$).
Because the exact operator is block diagonal under $P_H\oplus P_\perp$, this cross-branch sensitivity is consistent with representational entanglement rather than intrinsically harder harmonic dynamics.
All pressure correlations remain descriptive associations based on task-derived coordinates, not causal measurements of a model's internal mechanism.

\paragraph{Subspace leakage and error allocation.}
Figure~\ref{fig:extended-topology-audit}(c) uses pure non-harmonic inputs and measures leakage into the harmonic output,
\begin{equation}
L_{\perp\to H}=\|P_H\widehat S_TP_\perp w\|_M.
\end{equation}
The left panel reports total leakage, which generally grows with harmonic dimension $\beta_k$; the right panel divides the same quantity by $\sqrt{\beta_k}$.
The flatter normalized curves indicate that much of the raw increase is consistent with accumulation over more harmonic directions, rather than a uniform worsening of leakage per available direction.

For conventional mixed OOD inputs, Figure~\ref{fig:extended-topology-audit}(d) reports the fraction of squared prediction-error energy in the harmonic branch,
\begin{equation}
s_H(e)=\frac{\|P_He\|_M^2}{\|P_He\|_M^2+\|P_\perp e\|_M^2},
\qquad e=\widehat w_T-w_T.
\end{equation}
This is an error-allocation statistic, not a total-accuracy score: a high value means that a model's existing error is concentrated in the harmonic component.
GAOT assigns 50.6--95.9\% of OOD error energy to the harmonic branch across these six tasks, whereas TNO assigns only 0.2--1.2\%, illustrating the architecture dependence of where errors occur.

\paragraph{Model-side topology audit.}
We additionally apply inference-only controlled probes to the six tasks with nontrivial harmonic support.
For a unit-mass pure harmonic input $w_H$, the exact solution is $w_H$, giving the identity error $I_H=\|\widehat S_Tw_H-w_H\|_M$ and harmonic-to-non-harmonic leakage $L_{H\to\perp}=\|P_\perp\widehat S_TP_Hw\|_M$.
A pure non-harmonic input gives $L_{\perp\to H}$ above, while $E_\perp^{\mathrm{mix}}$ measures nonharmonic component error on the conventional mixed-input evaluation.
The three controlled-probe statistics use 286 geometry-level records from existing seed-0 checkpoints; all model--task--split cells use four geometries except RIGNO D-$k=2$ OOD, which uses two.
By contrast, $E_\perp^{\mathrm{mix}}$ is computed from the standard three-seed-median evaluation records, with 800 mixed-input records per model--task--split cell.
The controlled probes are structural stress tests rather than replacements for this conventional ranking.

\begin{table}[H]
\centering\scriptsize
\caption{Model-level topology audit. All errors are OOD medians; lower is better.}
\label{tab:model-topology-audit}
\begin{tabular}{@{}lrrrr@{}}
\toprule
Model & $I_H$ & $L_{H\to\perp}$ & $L_{\perp\to H}$ & $E_\perp^{\mathrm{mix}}$\\\midrule
MGN&.187&.131&.025&.848\\
RIGNO&.165&.114&.016&.894\\
Transolver&.329&.171&.012&1.137\\
GNOT&.213&.146&.014&1.040\\
GAOT&.588&.049&.005&.709\\
TNO&.349&.290&.004&.256\\
\bottomrule
\end{tabular}
\end{table}

The four empirical axes are not interchangeable.
TNO has the lowest mixed-input nonharmonic error ($E_\perp^{\mathrm{mix}}=0.256$) and very small $L_{\perp\to H}=0.004$, but its pure-harmonic identity error is $I_H=0.349$ and its reverse leakage is $L_{H\to\perp}=0.290$.
Conversely, GAOT has the smallest $L_{H\to\perp}=0.049$ but the largest $I_H=0.588$, showing that low cross-subspace leakage does not prevent attenuation or distortion within the harmonic subspace itself.
RIGNO attains the lowest empirical $I_H=0.165$, indicating the strongest harmonic-input preservation among the six evaluated models under this controlled probe.

\FloatBarrier
\section{Additional Limitations}
\label{app:limitations}

\paragraph{Topology and geometry scope.}
TopoBox-3D stratifies domains by numerically verified Betti vectors, but this is a homological notion of topology rather than a complete domain classification.
Betti numbers do not distinguish the fundamental group, knotting, torsion, or other embedding-dependent structure.
Moreover, the generator is restricted to connected box-minus-void domains assembled from disjoint primitive tunnels and cavities.
The conclusions therefore concern extrapolation beyond observed Betti support within this synthetic family, not arbitrary topological variation in CAD, anatomical, porous, or multiscale domains.

\paragraph{Operator and boundary-condition scope.}
The benchmark uses a linear, autonomous Hodge heat equation with fixed $\kappa=1$, evaluation time $T=0.1$, and homogeneous absolute boundary conditions.
Relative, mixed, interface, or physically imposed boundary conditions change the relevant cohomology, harmonic representatives, and boundary interactions.
Likewise, the present evidence does not establish the same behavior for forced, nonlinear, coupled, history-dependent, time-varying, or multiphysics operators.

\paragraph{Attribution and model-comparison scope.}
Protocols B--D necessarily change Betti support together with boundary count, mesh size, discretization statistics, and the positive spectrum.
Their matched IID--OOD ratios are therefore stress-test contrasts rather than causal effects of an individual Betti number.
The architecture comparison is also not a single-factor ablation: processors, adapters, native inputs, and inductive biases differ, TNO additionally receives incidence maps and a harmonic basis, and our TNO is a reimplementation because official code was unavailable.
Approximately matched parameter counts and identical training budgets do not imply that all six architectures are equally optimized.
Three seeds support the reported median and rank-level patterns but are insufficient for fine distinctions between neighboring models.

\paragraph{Diagnostic and statistical scope.}
The error-driver results are associative correlations and conditional rank regressions; they do not show that a network internally implements the corresponding spectral mechanism, and the target-retention statistic $q_T$ is partly coupled to the definition of relative error.
The topology-pressure summaries use only 12 task regimes, while the 36 model--task component cells repeat task-level pressure coordinates across models, so nominal correlation $p$-values do not replace replication over independent benchmark families.
In addition, $\Pi_H$ is nearly binary in TopoBox-3D and has not been validated as a graded topology-complexity measure.
Finally, the controlled subspace probes contain 286 geometry-level records from seed-0 checkpoints, with four geometries per model--task--split cell except for the two-geometry RIGNO D-$k=2$ OOD cell.
They reveal consistent empirical deviations from the exact subspace identities within this probe set, but do not carry the multi-seed uncertainty of the conventional benchmark; $E_\perp^{\mathrm{mix}}$ is reported separately from the full conventional mixed-input evaluation and is not interpreted as a pure-probe operator error.

\end{document}